\documentclass[11pt]{article}

\usepackage[colorlinks, linkcolor=blue, citecolor=blue]{hyperref}           
\usepackage{color}
\usepackage{graphicx,subfigure,amsmath,amssymb,amsfonts,bm,epsfig,epsf,url,dsfont}
\usepackage{amsthm}
\usepackage{tikz}
\usepackage{bbm}      
\usepackage{booktabs}
\usepackage{cases}
\usepackage{fullpage}
\usepackage[small,bf]{caption}
\usepackage[top=1in,bottom=1in,left=1in,right=1in]{geometry}
\usepackage{fancybox}

\usepackage{algorithm}
\usepackage{verbatim}
\usepackage{algorithmic}

\usepackage{mathtools}

\usepackage{scalerel,stackengine}
\stackMath
\newcommand\reallywidehat[1]{%
\savestack{\tmpbox}{\stretchto{%
  \scaleto{%
    \scalerel*[\widthof{\ensuremath{#1}}]{\kern-.6pt\bigwedge\kern-.6pt}%
    {\rule[-\textheight/2]{1ex}{\textheight}}%WIDTH-LIMITED BIG WEDGE
  }{\textheight}% 
}{0.5ex}}%
\stackon[1pt]{#1}{\tmpbox}%
}

\newcommand{\bE}{\mathbb{E}}

\newtheorem{theorem}{Theorem}

\newenvironment{fminipage}%
  {\begin{Sbox}\begin{minipage}}%
  {\end{minipage}\end{Sbox}\fbox{\TheSbox}}

\makeatletter
\newcommand*{\rom}[1]{\expandafter\@slowromancap\romannumeral #1@}
\makeatother

\newcommand{\Ind}{\mathds{1}}

\newcommand{\abs}[1]{\left|#1\right|}

\newcommand {\pr} {\mathbb{P}}

\newcommand{\calA}{{\cal A}}
\newcommand{\calB}{{\cal B}}

\newcommand{\calE}{{\cal E}}
\newcommand{\calF}{{\cal F}}

\newcommand{\calP}{{\cal P}}
\newcommand{\calQ}{{\cal Q}}

\newcommand{\calV}{{\cal V}}

\newcommand{\be}{\begin{equation}}
\newcommand{\ee}{\end{equation}}
\newcommand{\beqna}{\begin{eqnarray}}
\newcommand{\eeqna}{\end{eqnarray}}

\newcommand{\p}[1]{\left(#1\right)}
\newcommand{\pp}[1]{\left[#1\right]}
\newcommand{\ppp}[1]{\left\{#1\right\}}
\newcommand{\norm}[1]{\left\|#1\right\|}

\usepackage{xspace}
\newcommand\nth{\textsuperscript{th}\xspace}
\newcommand{\revt}{\mathsf{Rev}_t}
\newcommand{\sN}{\mathsf{N}}
\newcommand{\sV}{\mathsf{V_N}}

\begin{document}

\title{Sharp Thresholds of the Information Cascade Fragility Under a Mismatched Model}

\author{Wasim~Huleihel\thanks{W. Huleihel and O. Shayevitz are with the Department of Electrical Engineering-Systems at Tel-Aviv university, {T}el-{A}viv 6997801, Israel (e-mail:  \texttt{\{wasimh,ofersha@mail,eng\}.tau.ac.il}). This research was supported by the European Research Council (ERC) under grant agreement 639573.} \and Ofer~ Shayevitz\footnotemark[1]}

\date{}

\maketitle

\begin{abstract}
We analyze a sequential decision making model in which decision makers (or, players) take their decisions based on their own private information as well as the actions of previous decision makers. Such decision making processes often lead to what is known as the \emph{information cascade} or \emph{herding} phenomenon. Specifically, a cascade develops when it seems rational for some players to abandon their own private information and imitate the actions of earlier players. The risk, however, is that if the initial decisions were wrong, then the whole cascade will be wrong. Nonetheless, information cascade are known to be fragile: there exists a sequence of \emph{revealing} probabilities $\{p_{\ell}\}_{\ell\geq1}$, such that if with probability $p_{\ell}$ player $\ell$ ignores the decisions of previous players, and rely on his private information only, then wrong cascades can be avoided. Previous related papers which study the fragility of information cascades always assume that the revealing probabilities are known to all players perfectly, which might be unrealistic in practice. Accordingly, in this paper we study a mismatch model where players believe that the revealing probabilities are $\{q_\ell\}_{\ell\in\mathbb{N}}$ when they truly are $\{p_\ell\}_{\ell\in\mathbb{N}}$, and study the effect of this mismatch on information cascades. We consider both adversarial and probabilistic sequential decision making models, and derive closed-form expressions for the optimal learning rates at which the error probability associated with a certain decision maker goes to zero. We prove several novel phase transitions in the behaviour of the asymptotic learning rate.
\end{abstract}

\section{Introduction}\label{sec:intro}
There are myriad economic and social scenarios where our decisions are influenced by the actions of others. For example, voters are inclined to vote in favor of what opinion polls predicts will win. Academic researchers choose to work on topics that are of broad and current interest. Fertility decisions, e.g., how many children to have, are known to be influenced by what other people in the same geographical location are doing. Opinions we hold, products we buy, and technologies we use, etc., are all potentially affected by our surroundings. The above is a non-exhaustive list of scenarios where our rational behaviour guides us to follow the actions of others despite the fact that these may contradict our own information. 
%It is then evident that in many scenarios it may be rational for us to follow the actions of others despite the fact that these may contradict our own information. 
This is exactly the situation where information cascades \cite{Banerjee92asimple,Sushil92} develop. 

To illustrate the way information cascades evolve we consider the following simple and classical herding experiment, proposed and studied in \cite{Anderson1,Anderson2} (see also \cite[Ch. 16]{David10}). In this experiment, we place an urn that contains three marbles in front of a bunch of players. The urn contains either one red marble and two blue marbles (\emph{majority blue}), or, two red marbles and one blue marble (\emph{majority red}). Players do not know whether the urn is majority blue or red, while both urns are equally likely to be chosen. In a successive manner, each player randomly draws a single marble from the urn, memorizes its color, and returns it to the urn, while not showing it to the other players. Then, each player in his turn publicly announce his guess for the urn majority color. The players guesses are based on both their own private draws as well as the actions/announcements of previous players. 

Next, we explain how the above experiment evolves. It is clear that the first two players will announce their private signals as their guesses. Indeed, the first player gets to see his own draw only, and thus his best guess for the urn majority color is the color he draw. The second player is aware of that, and thus, together with his own draw he gets to see two independent draws from the urn. Accordingly, if the colors agree, then the player announces this color; otherwise, there is a tie, and in such a case let us assume that the player follows his own draw. Therefore, it is clear again that the second player announces the color of the marble he draw. Continuing, the third player now see three independent draws from the urn and consequently his best decision is the majority color among these draws. 

The most important observation here is that the rational guess of any subsequent player may not reflect its own private information. For example, if the first two announced colors were blue, then the third player guess is blue \emph{irrespectively} of the color of the marble he picked. Evidently, due to the fact that his guess will not reveal any information about the urn to any subsequent player, every subsequent player will guess the urn to be majority blue. This is where an \emph{information cascade} developed: while no one is under the impression that every player draw a blue marble, since the first two guesses were blue, future rational guesses must be blue as well. To wit, an information cascade is a sequence of decisions where it is optimal for players to ignore their own private information and imitate the decisions of players ahead of them. The problem with information cascade is that they can be \emph{wrong}! Indeed, if for instance in the above example the urn is majority red, then everyone wrongly announced blue as their guesses. In fact, in the above experiment, it can be shown that with probability $1/5$, a wrong cascade develops, in which case, most players will guess the urn majority color wrongly.  

The experiment above illustrates that information cascades can be wrong because they rely on very little actual information--the actions of the first few players can determine the actions of all future players. Nonetheless, this hints that information cascades can also be fundamentally very fragile. Indeed, suppose that in the above experiment two consecutive players $\ell$ and $\ell+1$ draw red marbles, and they cheat (or, act irrationally) by announcing their marbles despite the fact that a majority-blue cascade was already developed (say, the first two players announced blue). Then, it is clear that the wrong cascade can be broken: player $\ell+2$ sees four \emph{informative} draws (two blues and two reds), and he will announce his own private signal as his guess. The conclusion is that infusion of new information can overturn/brake wrong information cascades even after they have persisted for a long time. 

Motivated by the above observation, we consider a simple sequential decision making model in which not all players are rational, but rather certain players act irrationally, by revealing their own private signal and discarding the actions of previous players. More specifically, we focus on a recent model proposed by \cite{Peres18}, where it is  assumed that the $\ell$\nth player is irrational/revealer with some probability $p_\ell$, and is rational/Bayesian with complementary probability. While players \emph{do not} know whether other players before them were rational or not, it is assumed in \cite{Peres18} nonetheless that players \emph{know} the revealing probabilities $\{p_\ell\}_{\ell\in\mathbb{N}}$. As mentioned in \cite{Peres18}, this model is prompted by both empirical laboratory experiments \cite{Anderson2,Huck00,Georg10}, as well as several theoretical reasons \cite{Bernardo01}. One of the intriguing questions here, is whether wrong information cascades are broken in the above model? Or, stated differently, do people eventually learn the correct action? As was shown in \cite{Peres18}, the answer to this question is positive. Namely, it can be shown that there exists a sequence of revealing probabilities $\{p_\ell\}_{\ell\in\mathbb{N}}$ such that learning occurs. In particular, the optimal policy minimizing the probability of error is for the $\ell$\nth player to reveal its private signal with probability $p_\ell = c/\ell$, which in turn implies a learning rate of $c'/\ell$, where $c$ and $c'$ are explicit constants. While these results are neat, as mentioned in \cite{Peres18} they rely heavily on the assumption that players are fully coordinated, i.e., they know the revealing probabilities exactly, which might be unrealistic in practice. This sets precisely the main goal of our paper: we aim to understand how this coordination affects information cascades. To this end, we introduce a mismatch model where players believe that the revealing probabilities are $\{q_\ell\}_{\ell\in\mathbb{N}}$ when they truly are $\{p_\ell\}_{\ell\in\mathbb{N}}$. We are interested in understanding whether asymptotic learning occur in this case? and if so, under what conditions and at what learning rate? In particular, what is the cost of this mismatch?

The above mismatch model might be relevant in many real-world applications, such as, rumor spreading over social networks, online movie rating, etc., where it is well-documented that human behavior is sometimes irrational (e.g., \cite{Kahneman1973}). Furthermore, it is well-known in the social learning literature that a fully rational model often places unreasonable computational demands on Bayesian players (e.g., \cite{mossel2017}), hence understanding the impact of simpler more efficient strategies is desirable. As we explain in the paper, this situation can be partially captured by our model, since our mismatch framework allow for a family of sub-optimal strategies parameterized by the mismatch sequence $\{q_\ell\}_{\ell\in\mathbb{N}}$. 

\noindent\textbf{Main Contributions.} The main contributions of this paper are as follows. We start by formulating a general adversarial/worst-case model where the placement of irrational players is \emph{arbitrary}, and not being governed by any probabilistic/statistical rule. We show that this kind of model is in fact too stringent leading to trivial results. Combined with \cite{Peres18}, this fact motivates us to study the more flexible probabilistic model described above. For this model we characterize the asymptotic learning rate exactly, which turns out to exhibit several novel interesting phase transitions. Specifically, we first show that for either too ``optimistic" or ``pessimistic" assumptions, i.e., $q_t = o(p_t)$ or $q_t = \omega(p_t)$, asymptotic learning \emph{does not} occur, namely, the error probability is high, and the total number of wrong decisions is significant. We then consider the case where $q_t = \Theta(p_t)$, and show that asymptotic learning occurs, but at a reduced rate which loosely speaking depends on the ratio between the $\ell_1$-norms of the matched and mismatched revealing probabilities. This is true, as long as the magnitude of this ratio is moderate laying between two thresholds, otherwise, asymptotic learning does not occur! 

\noindent\textbf{Related work.} Sequential decision making has been studied in various areas including politics, economics and computer science. In particular, the case where all players are Bayesian (i.e., $p_\ell=0,\forall\ell$) was considered in \cite{David10,Banerjee92asimple,Sushil92}. Notably, \cite{Sushil92} gives many interesting real-world examples from diverse fields, where information cascades develop and are fragile. Similarly to \cite{Peres18}, our paper provides a more detailed theoretical study of the fragility phenomenon. In practice, however, it is well documented, that human behavior deviates from rationality, and rather irrational decisions are made often, see, e.g., \cite{Kahneman1973,Huck00} and \cite{Georg10}. Indeed, several laboratory experiments in \cite{Anderson1} illustrate that many individuals act irrationally, by ignoring the actions of other individuals and relying mainly on their own private information. Our model captures this empirically observed behavioral phenomenon. 

In \cite{Bernardo01}, a model which combines both rational and partially irrational (who put more weight on their private signal) types of individuals, as in our model, was studied. It is assumed, however, that players know which of the previous players were revealers. Using simulations it was suggested that learning is achievable only when completely irrational individuals exist, and that their fraction should vanish. These observations were rigorously proved in \cite{Peres18}, showing that the optimal number of revealers is logarithmic in the size of the group. Our results show that in many cases  learning is possible even if there is a mismatch. Recently, \cite{Tamuz18} also considered a model with irrational players, but assume that each agent knows which agents were revealers. While they show that asymptotic learning occurs, the optimal learning rate was not characterized, which is another contribution of our paper when there is a mismatch. Sequential decision models with unbounded private signals (e.g., Gaussian) were studied in \cite{Smith96,Caruthers17}. We also mention the study of sequential decision making models over social random graphs, e.g., \cite{Daron11,Anunrojwong18}. Finally, we mention another somewhat related literature which studies the situation where agents take repeated decisions (rather than just a single decision) based on their own private information and the actions of others, e.g., \cite{Matan18}, and one is interested in understanding whether all decision makers learn the correct state eventually, and if so at what speed. 

\section{Problem Setup}\label{sec:problem_setup}

In this section, we present our model and formulate the problem of interest. To convey neatly the main ideas of this paper, we focus on a simple setting of the information cascade model. Nonetheless, several generalizations listed at the end of this paper, can be derived using the same techniques used in this paper. Let $\theta\in\{1,2\}$ denote the state of the world, chosen uniformly at random. At times $t=1,2,3,\ldots$, players one by one try to guess $\theta$, relying on their own \emph{private} signals, as well as the \emph{global actions} (guesses) of players who played before them. 

We next describe the way private signals are formed. There is an urn that contains two types of marbles: Type--I marbles are blue, and Type--II are red. There are two hypotheses depending on the value of $\theta$. Specifically, given $\theta$, there are $\alpha$ marbles of type $\theta$ in the urn, and $\beta$ marbles of the other type, where we assume that $\alpha>\beta>0$. We conduct the following experiment: each player draws a single marble from the urn and replace it. The color he draw is his private signal. Thus, the private signals denoted by $X_1,X_2,\ldots$ are i.i.d., and:
\begin{align}
&\pr_1(X_t=1) = \frac{\alpha}{\alpha+\beta};\quad\quad \pr_1(X_t=2) = \frac{\beta}{\alpha+\beta},\nonumber\\
&\pr_2(X_t=1) = \frac{\beta}{\alpha+\beta};\quad\quad \pr_2(X_t=2) = \frac{\alpha}{\alpha+\beta},\nonumber
\end{align}
where $\pr_i(\cdot)\triangleq\pr(\cdot\vert\theta=i)$, and $t\in\mathbb{N}$. It is clear that the ratio $\alpha/\beta$, rather than the actual values of $\alpha$ and $\beta$, is important. We denote this ratio by $\gamma\triangleq\alpha/\beta$. Note that an alternative equivalent description of the above setting is that at each time $t$, the $t$\nth player private signal is $X_t=\theta$ with some probability, and $X_t=3-\theta$, with the complementary probability. With $\theta$ being latent, the players goal is to guess the type of the marbles in the urn correctly. We denote by $Z_t$ the guess of the $t$\nth player. As mentioned in the introduction, if all players act rationally by announcing their majority decision, then with positive probability (wrong) information cascade will occur. Accordingly, to break this wrong information cascade, we assume that each player, can operate in either one of the following two modes:
\begin{itemize}
\item A revealer/irrational player, whose guess is simply its private signal, i.e., $Z_t = X_t$.
\item A Bayesian/rational player, whose guess is its best estimate given his private signal, previous guesses by other players, and any additional auxiliary information.
\end{itemize}
Given the two modes above, it is left to specify the \emph{way} players are ``chosen" to be in either one of the above modes, which is the last aspect of our model. We start with a general \emph{adversarial/worst-case} machinery, which turns out to be too stringent, and as a consequence leads to trivial results. Nonetheless, this model will motivate our second \emph{probabilistic} setting, which is the focus of this paper.

In the adversarial setting, we assume that out of a total of $\mathsf{N}\in\mathbb{N}$ players $\mathsf{V_N}\in\mathbb{N}$ are irrational, and are chosen in an \emph{arbitrary} manner. We let $\Pi_\mathsf{N}$ be the set of these irrational players. Players \emph{do not} know whether previous players were rational or not, but they do have the value of $\mathsf{V_N}$ in advance. We define the probability of incorrect decision of the $t$\nth player, assuming that he is rational, as follows,
\begin{align}
\mathsf{P}_{\mathsf{adv},t}(\mathsf{V_N})&\triangleq\inf_{\hat{\theta}_t\in\hat{\Theta}}\sup_{\Pi_\mathsf{N}\subset[\mathsf{N}]:\;|\Pi_\mathsf{N}|=\mathsf{V_N}}\mathsf{P}_{e,t}(\hat{\theta}_t,\Pi_\mathsf{N}),\label{eqn:worst-case_error_def}
\end{align}
and
\begin{align}
\mathsf{P}_{e,t}(\hat{\theta}_t,\Pi_\mathsf{N})\triangleq\pr\pp{\left.\hat{\theta}_t(Z_1^{t-1},X_{t})\neq\theta\right|\mathsf{Rev}_\mathsf{N} =\Pi_\mathsf{N}},
\end{align}
where $Z_1^{t-1}$ is a shorthand notation for the sequence $(Z_1,Z_2,\ldots,Z_{t-1})$, $\mathsf{Rev}_{\mathsf{N}}$ designates the set of revealers, the maximum is taken over all possible sets of revealers of size $\mathsf{V_N}$, and the minimum is over the set of all possible estimators $\hat{\Theta}$, which are the Boolean maps $\{1,2\}^t\to\{1,2\}$. To wit, we look at the worst-case error probability over all possible choices of $\mathsf{V_N}$ irrational players out of $\mathsf{N}$ players. An alternative objective is to minimize the expected total number of errors, that is,
\begin{align}
\mathsf{TE}(\sV)\triangleq \inf_{\hat{\theta}\in\hat{\Theta}}\sup_{\Pi_\mathsf{N}\subset[\mathsf{N}]:\;|\Pi_\mathsf{N}|=\mathsf{V_N}}\sum_{t=1}^\sN\mathsf{P}_{e,t}(\hat{\theta}_t,\Pi_\mathsf{N}).\label{eqn:total_error_def}
\end{align}
For both objectives, it is a-priori unclear what is the optimal strategy $\hat{\theta}_t$. One option, which is simple and widely used in practice, is to assume that each rational player guesses the value of $\theta$ using the majority decision, denoted by $\mathsf{Maj}(Z_1^{t-1},X_{t})$. A more complicated approach is to minimize over the Boolean functions used by the rational players, i.e., to solve a minimax problem. We would like to find the asymptotic behaviour of $\mathsf{P}_{\mathsf{adv},t}(\mathsf{V_N})$ and $\mathsf{TE}(\sV)$, as a function of $\mathsf{V_N}$ and $t$. In particular, it is interesting to understand the structure of the worst-case choice of the set of the irrational players $\Pi_{\mathsf{N}}$. It turns out that the above model/objective, however, is too stringent. Specifically, we show in Section~\ref{app:too_string} that the error probability in \eqref{eqn:worst-case_error_def} associated with any estimator is lower bounded by a constant, and accordingly, the total number of errors in \eqref{eqn:total_error_def} is proportional to the number of players $\sN$. This implies that there are no guessing strategies that are \emph{robust} against an arbitrary adversarial revealers assignment. Therefore, a more flexible model is needed. 

To this end, we consider the probabilistic setting introduced in \cite{Peres18}. Here, we assume that the $t$\nth player is irrational with probability $p_{t}$, independently of the other players. Accordingly, this means that if player $t$ is irrational, then $Z_t=X_{t}$, while if he is rational, then $Z_t=\hat{\theta}_t(Z_1,\ldots,Z_{t-1},X_{t})$, where $\hat{\theta}$ is a certain estimator for $\theta$. The main important ingredient of our model is that we assume that players are completely oblivious to whether previous players were revealers or not. To wit, contrary to previous related works (e.g., \cite{Bernardo01,Tamuz18,Peres18}), we assume that revealers neither know the exact positioning of revealers, nor the underlying probabilistic law of their placements. Instead, players assume that other players can be revealers with probabilities $\calQ\equiv\{q_t\}_{t=1}^\infty$, which might be different than the actual underlying probabilities $\calP\equiv\{p_t\}_{t=1}^\infty$. In that case, we say that there is a mismatch. Thus, estimators $\hat{\theta}_t$ might be in fact a function of $\calQ$ as well. The matched case case where $\calP=\calQ$ was considered in \cite{Peres18}.

Whenever a player is rational we assume that he tries to do his best in guessing $\theta$ under the knowledge of $\calQ$. Namely, a rational player employs the (mismatched) maximum a posteriori probability (MAP) estimator, which is simply the MAP estimator, but with $\calP$ replaced by $\calQ$. Specifically, for $i\in\{1,2\}$ and $t\geq1$, define the distributions:
\begin{align}
W_i^t(z_1^t)&\triangleq\pr_i\p{Z_1^t=z_1^t}\label{notation:1},\\
H_i^t(z_1^{t-1},x_t)&\triangleq\pr_i\p{Z_1^{t-1}=z_1^{t-1},X_{t}=x_{t}},
\end{align}
and the corresponding likelihood r.v.s by
\begin{align}
\mathsf{L}_{i}^{t}&\triangleq W_i^t(Z_1^t),\\
\mathsf{D}_{i}^{t}&\triangleq H_i^t(Z_1^{t-1},X_t),
\end{align}
with $\mathsf{L}_{i}^{0}=\mathsf{D}_{i}^{0}=1$, for $i\in\{1,2\}$. Note that player $t$ can compute the likelihoods $\mathsf{D}_{i}^{t}$, for $i=1,2$. The likelihoods $\mathsf{L}_{i}^{0}=\mathsf{D}_{i}^{0}=1$, for $i\in\{1,2\}$, can be computed by a ``genie" who observes the decisions of the first $t$ players. The above probabilities and likelihoods are certain quite complicated functions of $\calP$. We let $\hat{W}_i^t$ and $\hat{H}_i^t$ be the corresponding probabilities with $\calP$ replaced by $\calQ$. We also define $\hat{\mathsf{L}}_{i}^{t}$ and $\hat{\mathsf{D}}_{i}^{t}$ in the same way, but with $W_i^t$ and $H_i^t$ replaced by $\hat{W}_i^t$ and $\hat{H}_i^t$, respectively. With these definitions, the mismatched MAP estimate, denoted by $Z_t=\mathsf{MAP}_{\calQ}(Z_1^{t-1},X_{t})$, is:
\begin{align}
    \mathsf{MAP}_{\calQ}(Z_1^{t-1},X_{t}) \triangleq \begin{cases}
    1,\ &\hat{\mathsf{D}}_{1}^{t}>\hat{\mathsf{D}}_{2}^{t},\\
    2,\ &\hat{\mathsf{D}}_{1}^{t}<\hat{\mathsf{D}}_{2}^{t},\\
    X_t,\ &\hat{\mathsf{D}}_{1}^{t}=\hat{\mathsf{D}}_{2}^{t}.
    \end{cases}\label{notation:5}
\end{align}
Thus, the mismatch aspect in our setup lies in the fact that the $t$\nth player estimator function depends on $\calQ$, and more importantly, independent of $\calP$. We mention here that the above estimator in fact provides a family of possible estimators indexed by $\calQ$. For instance, $\calQ\equiv1$ corresponds to majority decisions, for which rational players guesses are simply the majority color of their own private and previously announced signals. 

Players try to guess the urn majority color. We define the probability of incorrect guess by the $t$\nth player, as follows,
\begin{align}
\mathsf{P}_{e,t}(\calP,\calQ)\triangleq\pr\p{Z_t\neq\theta}.
\end{align}
Our goal is to understand the asymptotic learning rate at which the above error probability decays to zero as a function of $t$ (i.e., learning occurs). To motivate and define our objective precisely, we recall the following recent result which deals with the matched case where $\calP=\calQ$.
\begin{theorem}{\cite[Theorem 1.1]{Peres18}}\label{thm:1}
Let 
\begin{align}
\kappa(\gamma)\triangleq\pp{1+\frac{\gamma-1}{\log\gamma}\p{\log\frac{\gamma-1}{\log\gamma}-1}}^{-1}.\label{eqn:spe_const}
\end{align}
Then,
\begin{align}
\inf_{\calP}\limsup_{t\to\infty}t\cdot\mathsf{P}_{e,t}(\calP,\calP) = \kappa(\gamma).
\end{align}
Moreover, one can be arbitrarily close to the optimum by taking,
\begin{align}
p_t^\star = (1+\varepsilon)\cdot\frac{(1+\gamma)\kappa(\gamma)}{t}\wedge 1,\label{eqn:optimal_distribution_matched}
\end{align}
for $t\geq1$ and arbitrary $\epsilon>0$. 
\end{theorem}
Theorem~\ref{thm:1} states that the optimal learning rate is $\Theta(1/t)$, and furthermore provides the exact leading constant in \eqref{eqn:spe_const}. To achieve this optimal learning rate, the revealing probabilities should also decay as $\Theta(1/t)$. The intuitive reasoning behind these findings can be found in \cite[Sec. 1.2]{Peres18}. With this result in mind, in the mismatch case where $\calQ\neq\calP$, we focus on the following scenario. We assume that $\calP = \calP^\star$, where $\calP^\star$ is defined in \eqref{eqn:optimal_distribution_matched}. In other words, the underlying revealing probabilities sequence is the \emph{optimal} one, while players assume a (possibly) \emph{different} sequence of revealing probabilities $\calQ$. For simplicity of demonstration, we opted to focus on this special case since we found it to be the most natural one. Nonetheless, our techniques apply also for the more general case where $\calP\neq\calP^\star$, and at the end of the following section we cover with this case too. It is then interesting to understand whether such a mismatch has any effect whatsoever on the achieved learning rate. In particular, does asymptotic learning always occur? or, perhaps there is a sequence of revealing probabilities $\calQ$ for which learning is impossible. To answer these questions, we aim to characterize the \emph{polynomial learning rate}, defined as follows,
\begin{align}
\mathsf{E}(\calP,\calQ)\triangleq\liminf_{t\to\infty}-\frac{\log\mathsf{P}_{e,t}(\calP,\calQ)}{\log t}.
\end{align}
To lower bound $\mathsf{E}(\calP,\calQ)$ we upper bound the error probability $\mathsf{P}_{e,t}(\calP,\calQ)$, which in turn can be used to upper bound the expected total number of errors: 
\begin{align}
 \mathsf{TE}_t\triangleq\bE\pp{\sum_{\ell=1}^t\Ind\pp{Z_{\ell}\neq\theta}}.
\end{align}
Accordingly, a positive polynomial decaying learning rate implies that $\mathsf{TE}_t = o(t)$, i.e., the number of errors is negligible compared to the total number of players participated thus far. It is clear that Theorem~\ref{thm:1} implies that $\mathsf{E}(\calP^\star,\calP^\star)=1$, and in fact that $0\leq\mathsf{E}(\calP^\star,\calQ)\leq1$, for any $\calQ$. We would like to understand when it is possible or impossible to obtain a positive polynomial decaying learning rate, i.e., when $\mathsf{E}(\calP^\star,\calQ)>0$. Note that an interesting question is to characterize the specific constant in front of the polynomial decaying term by evaluating $\lim_{t\to\infty}t^{\mathsf{E}(\calP,\calQ)}\cdot\mathsf{P}_{e,t}(\calP,\calQ)$, which we leave as an open question for future research. Finally, note that Theorem~\ref{thm:1} gives a simple lower bound on $\mathsf{P}_{e,t}(\calP^\star,\calQ)$ because of the trivial inequality $\mathsf{P}_{e,t}(\calP^\star,\calQ) \geq\mathsf{P}_{e,t}(\calP^\star,\calP^\star)$. Indeed, for \emph{each} player seeking to minimize the error probability, its best action is to output the (matched) MAP decision. It is interesting to note that this observation also follows from
\begin{align}
\mathsf{P}_{e,t}(\calP^\star,\calQ) \geq  \frac{\beta}{\alpha+\beta}\cdot p_t^\star= \frac{\kappa(\gamma)}{t},
\end{align}
for $t$ large enough, and the first inequality follows because $\frac{\beta}{\alpha+\beta} p_t^\star$ is the error probability when player $t$ acts on its private information only (i.e., revealer), while we ignore the error resulted when player $t$ is rational. This lower bound, however, is not tight as we show in the following section. For the rest of this paper, we let $\calP_t\triangleq (p_1,p_2,\ldots,p_t)$, and $\norm{\cdot}$ denotes the $\ell_1$-norm. 

\section{Main Results}\label{sec:main}

In this section, we study the probabilistic setting described in the previous section. According to Theorem~\ref{thm:1}, to achieve the optimal learning rate the revealing probabilities should decay as $\Theta(t^{-1})$. Note that it is clear that the revealing probabilities cannot decay to zero too quickly. Indeed, if for example $\norm{\calP}<\infty$, then by the Borel-Cantelli lemma there will be only a finite number of revealers almost surely, which is equivalent to the situation where no revealers exist. This in turn implies a non-vanishing error probability. The situation is somewhat similar when the revealing probabilities decay to zero too slowly. Intuitively, in this case, it can be shown that the error probability is dominated by the probability that the $t$\nth player is a revealer and announce a wrong decision, namely, $\frac{\beta}{\alpha+\beta}p_t$. Therefore, if for example $p_t = \Theta(t^{-c})$, for some $c\in(0,1)$, then $\mathsf{E}(\calP,\calP) = c< \mathsf{E}(\calP^\star,\calP^\star)$. 

The optimal scaling of the revealing probabilities suggests that with high probability there should be $\norm{\calP_t}\sim\log t$ revealers, as $t\to\infty$. Accordingly, in terms of mismatch, it makes sense that ``wise" players will assume that the revealing probabilities decay at the same order, but perhaps with a different constant in front, e.g., $q_t = \rho\cdot p_t$, and $\rho\neq1$. Nonetheless, as mentioned above, rather than modeling the imperfect knowledge of players about the revealing probabilities, our mismatched MAP can also model the situation where players intentionally employ sub-optimal strategies (e.g., in order to reduce computational complexity), such as when $\calQ\equiv1$, which results in majority decisions. While it is clear from the above that in the matched case it is strictly worse to assume that the revealing probabilities $\calP$ decay to zero too quickly/slowly, it is a-priori unclear if this is true for $\calQ$ as well. The following result shows that if the \emph{assumed} revealing probabilities $\calQ$ are either too small or too large, then asymptotic learning does \emph{not} occur at all! In particular, $\mathsf{E}(\calP^\star,\calQ)=0$. We have the following result.
\begin{theorem}[Too quick/slow]\label{thm:2}
For any sequence of mismatched revealing probabilities $\calQ$ such that,
\begin{align}
\liminf_{t\to\infty}\frac{\norm{\calQ_t}}{\log t}=0\quad\mathrm{or}\quad
\limsup_{t\to\infty}\frac{\norm{\calQ_t}}{\log t}=\infty,
\end{align}
we have $\mathsf{E}(\calP^\star,\calQ)=0$.
\end{theorem}
Theorem~\ref{thm:2} implies, for example, that if $q_t =o(t^{-1})$ or $q_t =\omega(t^{-1})$, then the error probability cannot decay to zero polynomially fast. In fact, our proof gives a general lower bound on the probability of error which implies for instance that when $\norm{\calQ}<\infty$, then the probability of error is lower bounded by a constant. Indeed, as will be seen in the proofs, to analyze the error probability one needs to track the dynamics of the likelihood ratio $\mathsf{R}_t\triangleq\frac{\mathsf{L}_1^t}{\mathsf{L}_2^t}$. In particular, $\mathsf{R}_t<\beta/\alpha$ ($\mathsf{R}_t>\alpha/\beta$) implies that the $t$\nth player MAP decision is ``$2$" (``$1$"). Accordingly, given that $\theta=1$, the main observation in the proof of Theorem~\ref{thm:2}, is based on the realization that when $\norm{\calQ_t} = O(1)$, even in the worst-case scenario where the majority of the decisions before player $t$ were $\hat{\theta}_i=1$, only a \emph{finite} number of wrong decisions (namely, $\hat{\theta}_i=2$) suffice to mislead player $t$ and output $\hat{\theta}_t=2$. This happens to be the case because of the fact that the likelihood ratio depends on $\calQ$ only through $\norm{\calQ}$, and therefore, it cannot diverge. The intuitive explanation for the obtained result when $\norm{\calQ_t}\gg\log t$ is given after Theorem~\ref{thm:3}.  

We next consider the more interesting case where $q_t = \rho\cdot p_t$, for $t\in\mathbb{N}$ and $\rho\in\mathbb{R}_+$, or, more generally, $\norm{\calQ_t}/\norm{\calP^\star_t}\to\rho$, as $t\to\infty$. To present our main result we establish first some notation. Let 
\begin{align}
\rho_0&\triangleq \frac{\log\gamma}{\gamma-1}\label{eqn:rhodef},%\\
%\rho_1&\triangleq \frac{\gamma\cdot\log\gamma}{\gamma-1}.
\end{align}
and $\rho_1\triangleq\gamma\cdot\rho_0$. Also, define
%\begin{align}
%\resizebox{0.91\columnwidth}!{$\delta(\gamma,\rho)\triangleq \frac{\gamma\log\gamma-\rho(\gamma-1)}{(1+\gamma)\log\gamma}-\rho\frac{\gamma-1}{\gamma+1}\pp{1-\frac{\log\frac{\rho(\gamma-1)}{\log\gamma}}{\log\gamma}}.\label{eqn:delta_def}$}
%\end{align}
\begin{align}
\delta(\gamma,\rho)&\triangleq \frac{\gamma\log\gamma-\rho(\gamma-1)\pp{1+\log\frac{\gamma\log\gamma}{\rho(\gamma-1)}}}{(1+\gamma)\log\gamma}.\label{eqn:delta_def}
\end{align}
\begin{theorem}[Multiplicative mismatch]\label{thm:3}
For any sequence of mismatched revealing probabilities $\calQ$ such that,
\begin{align}
\limsup_{t\to\infty}\frac{\norm{\calQ_t}}{\norm{\calP^\star_t}}=\rho\in\mathbb{R}_+,
\end{align}
we have:
\begin{itemize}
\item If $\rho\leq\rho_0$ or $\rho\geq\rho_1$,
$$\mathsf{E}(\calP^\star,\calQ)=0.$$
\item If $\rho_0\leq\rho\leq1$,
$$\mathsf{E}(\calP^\star,\calQ)=(1+\gamma)\pp{\delta(\gamma,\rho)-\frac{\gamma-1}{\gamma+1}(1-\rho)}\kappa(\gamma).$$
\item If $1\leq\rho\leq\rho_1$,
$$\mathsf{E}(\calP^\star,\calQ)=(1+\gamma)\delta(\gamma,\rho)\kappa(\gamma).$$
\end{itemize}
\end{theorem}
\begin{figure}[!t]
\begin{minipage}[b]{1.0\linewidth}
  \centering
	\centerline{\includegraphics[width=8.5cm,height = 6.5cm]{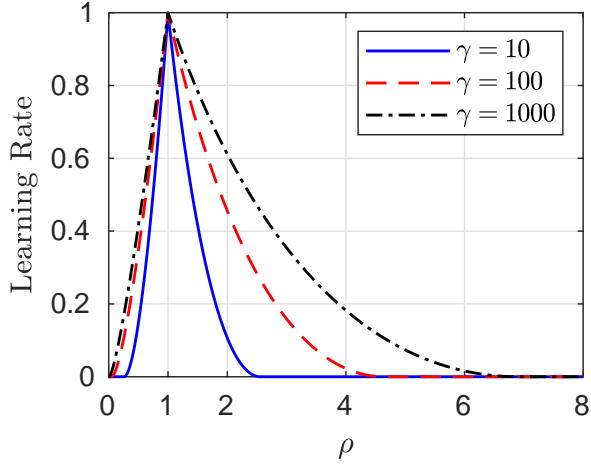}}
	\end{minipage}
	\caption{The learning rate $\mathsf{E}(\calP^\star,\calQ)$ in Theorem~\ref{thm:3} as a function of $\rho$, for different values of $\gamma$. The maximal learning rate is achieved at $\rho=1$ for any $\gamma$, as expected. Also, as $\gamma$ increases the range of $\rho$'s for which the learning rate is positive increases.}
	\label{fig:1}
\end{figure}
It can be checked that $\mathsf{E}(\calP^\star,\calQ)$ is continuous in $(\rho,\gamma)$. Theorem~\ref{thm:3} suggests a clear phase transition in the behaviour of the learning rate (see Fig.~\ref{fig:1} for a numerical illustration). To wit, even when players assume that the revealing probabilities decay at the same order of the optimal revealing assignments, albeit with a different constant, there are regimes where the learning rate is zero (i.e., when $\rho\leq\rho_0$ and $\rho\geq\rho_1$). We next give a heuristic explanation as to why the learning rate is zero in these regimes. We start with the case where $\rho\geq\rho_1$. First, note that there are two main sources for wrong action: 1) the $t$\nth player is irrational, which happens with probability $p_t^\star$, and his draw is of minority color type, or, 2) the $t$\nth player is rational, which happens with probability $1-p_t^\star$, but his mismatched MAP estimate is wrong. Therefore, it is clear that the error probability can be lower bounded by $\mathsf{P}_{e,t}(\calP^\star,\calQ)\geq(1-p^\star_t)\cdot\pr_1\p{\mathsf{MAP}_{\calQ}(Z_1^{t-1},X_{t})=2}$. We show in the proof that the MAP error probability can be further lower bounded by the probability that the likelihood ratio $\mathsf{R}_{t-1}\triangleq\mathsf{L}_1^{t-1}/\mathsf{L}_2^{t-1}$, at time $t-1$, is less than $\beta/\alpha$, namely, we have $\mathsf{P}_{e,t}(\calP^\star,\calQ)\geq(1-p_t^\star)\cdot\pr_1\p{\mathsf{R}_{t-1}<\beta/\alpha}$. To further lower bound the probability term on the r.h.s. of the above inequality, we construct a particular trajectory that ensures that the likelihood ratio $\mathsf{R}_{t-1}$ stays always below the threshold $\beta/\alpha$. The main observation is then that when the likelihood ratio is bellow $\beta/\alpha$, the corresponding log-likelihood ratio process $\log\mathsf{R}_t$ as a function of $t$, performs a random walk with a \emph{downward} drift \emph{when} $\rho\geq\rho_1$, and thus intuitively the probability that the likelihood ratio will stay bellow some fixed value is high. Technically speaking, we show that this random walk is a supermartingale, and by using well-known tail probability bounds for such processes, we show that $\pr_1\p{\mathsf{R}_{t-1}<\beta/\alpha}$ can not decay too fast. Establishing this result, and proving a certain monotonicity property of the error probability w.r.t. $\calQ$, we show that $\mathsf{E}(\calP^\star,\calQ)=0$ when $\norm{\calQ_t}\gg\log t$ as well, as claimed in Theorem~\ref{thm:2}. On other hand, when $\rho<\rho_1$, the previously mentioned random walk has an upward drift, and thus, the probability that this walk stays always bellow $\beta/\alpha$ is intuitively small, and in fact, decays at the polynomial rate given in Theorem~\ref{thm:3}. 

The reason for the learning rate being zero for $\rho\leq\rho_0$ is similar. Contrary to the case where $\rho\geq\rho_1$, in this regime, it can be shown that bellow $\log(\beta/\alpha)$ the log-likelihood ratio process performs a random walk with an upward drift, and thus the approach used before for lower bounding the error probability is not going to work. It turns out, however, that above $\log(\alpha/\beta)$, the log-likelihood ratio process has a downward drift. Moreover, we can show that in this case the walk cannot diverge, or, more precisely, go beyond a certain \emph{finite} value. Among other things, this implies that it takes only a finite number of timestamps to drive the log-likelihood ratio bellow $\log(\beta/\alpha)$, which in turn entails that the error probability is finite as well. Specifically, to lower bound the error probability, we show that it is suffice to look at all trajectories for which the private signals of the last $t^\star\in\mathbb{N}$ revealers are opposite to the majority (e.g., $X_i=2$, for if $\theta=1$). Indeed, this way we can assure that the likelihood ratio decreases by a multiplicative factor of $\beta/\alpha$. Accordingly, since we argue that the maximal value that likelihood ratio can attain is finite, it is clear that there exists a finite value of $t^\star$ which will drive $\mathsf{R}_t$ bellow $\beta/\alpha$ (note that $t^\star$ revealers decrease the likelihood by an exponential factor of $(\beta/\alpha)^{t^\star}$). Finally, when $\rho_0\leq\rho\leq1$, the random walk has now an upward drift, and thus, the probability that it will go bellow $\beta/\alpha$ is small, and in fact, decaying at the polynomial rate given in Theorem~\ref{thm:3}. 

The above results characterize $\mathsf{E}(\calP^\star,\calQ)$. As mentioned in the previous section, the same techniques exactly can be used to derive the learning rate $\mathsf{E}(\calP,\calQ)$ for any $\calP$, and we present our main findings bellow. Proof sketches can be found in Section~\ref{app:additional_Proofs}. First, as was mentioned at the beginning of this section whenever $\calP$ is such that $\norm{\calP}<\infty$, then Borel-Cantelli lemma implies that there will be only a finite number of revealers almost surely, which is equivalent to the situation where no revealers exist. This in turn implies a non-vanishing trivial error probability, and asymptotic learning does not occur. In fact, if $\norm{\calP_t}=o(\log t)$, then $\mathsf{E}(\calP,\calQ)=0$, as we show in Section~\ref{app:additional_Proofs}. 
\begin{theorem}\label{thm:4}
For any sequence of revealing probabilities $\calP$ such that $\norm{\calP_t} = o(\log t)$, and any sequence of mismatched revealing probabilities $\calQ$, it holds that 
$\mathsf{E}(\calP,\calQ)=0$.
\end{theorem}
Next, we consider the case where $\norm{\calP_t}/\log t\to \mathsf{C_p}$, as $t\to\infty$, which happens to be the case, for example, when $p_t = \mathsf{C_p}/t\wedge 1$, for some $\mathsf{C_p}\in\mathbb{R}_+$. Recall the definitions of $\rho_0$, $\rho_1$, and $\delta(\gamma,\rho)$ in \eqref{eqn:rhodef}--\eqref{eqn:delta_def}. We have the following result.
\begin{theorem}\label{thm:5}
For any sequence of revealing probabilities $\calP$ such that $\norm{\calP_t}/\log t\to \mathsf{C_p}$, for some $\mathsf{C_p}\in\mathbb{R}_+$, and mismatched revealing probabilities $\calQ$ such that,
\begin{align}
\limsup_{t\to\infty}\frac{\norm{\calQ_t}}{\norm{\calP_t}}=\rho\in\mathbb{R}_+,
\end{align}
we have:
\begin{itemize}
\item If $\rho\leq\rho_0$ or $\rho\geq\rho_1$, then $$\mathsf{E}(\calP,\calQ)=0.$$
\item If $\rho_0\leq\rho\leq1$,
$$\mathsf{E}(\calP,\calQ)=1\wedge\pp{\mathsf{C_p}\cdot\p{\delta(\gamma,\rho)-\frac{\gamma-1}{\gamma+1}(1-\rho)}}.$$
\item If $1\leq\rho\leq\rho_1$, then $$\mathsf{E}(\calP,\calQ)=1\wedge\pp{\mathsf{C_p}\cdot\delta(\gamma,\rho)}.$$
\end{itemize}
\end{theorem}
\begin{figure}[!t]
\begin{minipage}[b]{1.0\linewidth}
  \centering
	\centerline{\includegraphics[width=8.5cm,height = 6.5cm]{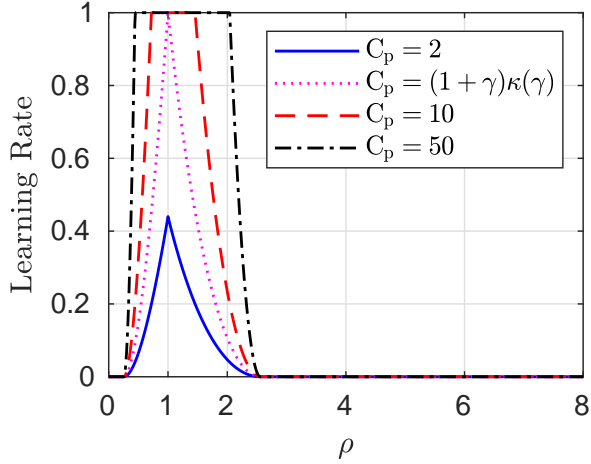}}
	\end{minipage}
	\caption{The learning rate $\mathsf{E}(\calP,\calQ)$ in Theorem~\ref{thm:5} as a function of $\rho$, for different values of $\mathsf{C_p}$, and $\gamma=10$.}
	\label{fig:2}
\end{figure}
From Theorem~\ref{thm:5} it can be seen that conceptually the learning rate behaves similarly to the learning rate when $\calP=\calP^\star$. In particular, the learning exhibits the same phase transitions as in Theorem~\ref{thm:3}. Fig.~\ref{fig:2} presents a numerical calculation of the rate in Theorem~\ref{thm:5}, for several values of $\mathsf{C_p}$. Note that for $\rho=1$, when $\mathsf{C_p}<(1+\gamma)\kappa(\gamma)$, we get that $\mathsf{E}(\calP,\calQ)<1$, while, for any $\mathsf{C_p}\geq(1+\gamma)\kappa(\gamma)$, we get that $\mathsf{E}(\calP,\calQ)=1$. This might seem counterintuitive because Theorem~\ref{thm:1} claims that $\mathsf{C_p}=(1+\gamma)\kappa(\gamma)$ is the optimal value minimizing the error probability, while the above suggests that any value $\mathsf{C_p}\geq(1+\gamma)\kappa(\gamma)$ suffices. Note, however, that while indeed any value $\mathsf{C_p}\geq(1+\gamma)\kappa(\gamma)$ gives a unit polynomial learning rate, the choice of $\mathsf{C_p}=(1+\gamma)\kappa(\gamma)$ minimizes the leading coefficient in front of the decaying term, namely, $\limsup_{t\to\infty}t\cdot\mathsf{P}_{e,t}(\calP^\star,\calP^\star)<\limsup_{t\to\infty}t\cdot\mathsf{P}_{e,t}(\calP,\calP)$, for any $\calP$ with $\mathsf{C_p}>(1+\gamma)\kappa(\gamma)$. This explains also why $\mathsf{E}(\calP,\calQ)$ in Theorem~\ref{thm:5} is increasing as a function of $\mathsf{C_p}$. Specifically, it is seen that when $\mathsf{C_p}>(1+\gamma)\kappa(\gamma)$, there are values of $\rho$ for which $\mathsf{E}(\calP^\star,\rho\cdot\calP^\star)<\mathsf{E}(\calP,\rho\cdot\calP)$. Indeed, in case of mismatch, taking $\calP^\star$ to be the underlying revealing probabilities might be sub-optimal, and choosing a different set of probabilities which combat the mismatch results in a higher rate. Finally, we consider the case where $\norm{\calP_t}=\omega(\log t)$, for which we have the following result.
\begin{theorem}\label{thm:6}
Let $\calP$ be such that $\norm{\calP_t}/\log t\to \infty$. If the mismatched revealing probabilities $\calQ$ is such that,
\begin{align}
\limsup_{t\to\infty}\frac{\norm{\calQ_t}}{\norm{\calP_t}}=\rho\in\mathbb{R}_+,
\end{align}
with $\rho\leq\rho_0$ or $\rho\geq\rho_1$, or
$$
\liminf_{t\to\infty}\frac{\norm{\calQ_t}}{\norm{\calP_t}}=0\quad\mathrm{or}\quad
\limsup_{t\to\infty}\frac{\norm{\calQ_t}}{\norm{\calP_t}}=\infty,
$$
then $\mathsf{E}(\calP,\calQ)=0$. Otherwise, if $\rho_0<\rho<\rho_1$, then
$$
\mathsf{E}(\calP,\calQ) = \lim_{t\to\infty}-\frac{\log p_t}{\log t}.
$$
\end{theorem}
Theorem~\ref{thm:6} states that if the number of revealers is significantly bigger than $\log t$, then the error probability is dominated by the probability that the $t$\nth player is a revealer and announce a wrong decision, namely, $\frac{\beta}{\alpha+\beta}p_t$. Therefore, if for example $p_t = \Theta(t^{-c})$, for some $c\in(0,1)$, then $\mathsf{E}(\calP,\calQ) = c$, as long as the mismatch is not too ``severe", namely, $\norm{\calQ_t}/\norm{\calP_t}\to\rho$, and $\rho_0<\rho<\rho_1$. Otherwise, learning does not occur and $\mathsf{E}(\calP,\calQ)=0$. 

%%%%%%%%%%%%%%%%%%%%%%%%%%%%%%%%%%%%%%%%%%%%%%%%%%%%%%%%

\section{Proof of Main Results}
\subsection{Proof of Theorem~\ref{thm:2}}
We start with the case where $\liminf_{t\to\infty}\frac{\norm{\calQ_t}}{\log t}=0$. To lower bound the error probability we need to understand first how the mismatched MAP decisions evolve over time/players. We next establish some notation, which will simplify the analysis. For $i\in\{1,2\}$ and $t\geq1$, recall our notation in \eqref{notation:1}--\eqref{notation:5}. 
Also, for $x\in\{1,2\}$, let
\begin{align}
\phi_i(x)\triangleq\frac{\alpha}{\alpha+\beta}\mathds{1}\pp{x=i}+\frac{\beta}{\alpha+\beta}\mathds{1}\pp{x\neq i}.\label{eqn:phi_def}
\end{align}
Note that $\hat{\mathsf{D}}_{i}^{t} = \hat{\mathsf{L}}_{i}^{t-1}\phi_i(X_{t})$. Finally, we define 
\begin{align}
\mathsf{R}_t\triangleq\frac{\hat{\mathsf{L}}_{1}^{t}}{\hat{\mathsf{L}}_{2}^{t}},\quad\quad\quad
\mathsf{R}'_t\triangleq\frac{\hat{\mathsf{D}}_{1}^{t}}{\hat{\mathsf{D}}_{2}^{t}}.
\end{align}
It is clear that 
\begin{align}
\mathsf{R}'_t = \mathsf{R}_{t-1}\frac{\phi_1(X_t)}{\phi_2(X_t)}.\label{eqn:Rtprimedef}
\end{align}
Since the ratio $\phi_1(X_t)/\phi_2(X_t)$ can take values in $\{\beta/\alpha,\alpha/\beta\}$, we have three possible cases only:
\begin{itemize}
\item If $\mathsf{R}_{t-1}<\beta/\alpha$ then, clearly $\mathsf{R}'_t<1$, irrespective of the value of $X_t$. Thus, $Z_t=1$ only if player $t$ is irrational and $X_t=1$. Otherwise, $Z_t=2$.
\item If $\mathsf{R}_{t-1}\in[\beta/\alpha,\alpha/\beta]$ then it can be easily shown that $Z_t=X_t$.
\item If $\mathsf{R}_{t-1}>\alpha/\beta$ then, clearly $\mathsf{R}'_t>1$, irrespective of the value of $X_t$. Thus, $Z_t=2$ only if player $t$ is irrational and $X_t=2$. Otherwise, $Z_t=1$.
\end{itemize}
Let $\ppp{\calF_t}_{t\geq0}$ denote the filtration spanned by $\ppp{Z_t}_{t\geq1}$. Let $\hat{\pr}_1(Z_t=i\vert\calF_{t-1})$ be the probability that the guess by player $t$ is $i$, given the history and $\theta=1$, and the evaluation of this probability is with respect to the mismatched revealers probabilities $\calQ$. Then, based on the above, we have
\begin{align}
\hat{\pr}_1(Z_t=1\vert\calF_{t-1})=\begin{cases}
\frac{\alpha}{\alpha+\beta}q_t,\ &\;\text{if}\quad\mathsf{R}_{t-1}<\beta/\alpha,\\
\frac{\alpha}{\alpha+\beta},\ &\;\text{if}\quad\mathsf{R}_{t-1}\in[\beta/\alpha,\alpha/\beta],\\
1-\frac{\beta}{\alpha+\beta}q_t,\ &\;\text{if}\quad\mathsf{R}_{t-1}>\alpha/\beta,
\end{cases}\label{eqn:mis_like1}
\end{align}
and
\begin{align}
\hat{\pr}_2(Z_t=1\vert\calF_{t-1})=
\begin{cases}
\frac{\beta}{\alpha+\beta}q_t,\ &\;\text{if}\quad\mathsf{R}_{t-1}<\beta/\alpha,\\
\frac{\beta}{\alpha+\beta},\ &\;\text{if}\quad\mathsf{R}_{t-1}\in[\beta/\alpha,\alpha/\beta],\\
1-\frac{\alpha}{\alpha+\beta}q_t,\ &\;\text{if}\quad\mathsf{R}_{t-1}>\alpha/\beta.
\end{cases}\label{eqn:mis_like2}
\end{align}
There are two main sources for wrong action: 1) the $t$\nth player is irrational, which happens with probability $p_t$, and his draw is of minority color type, or, 2) the $t$\nth player is rational, but his mismatched MAP estimate is wrong. Accordingly, we can write
\begin{align}
\mathsf{P}_{e,t}(\calP^\star,\calQ) &= \pr\p{\mathsf{MAP}_{\calQ}(Z_1^{t-1},X_{t})\neq\theta}\cdot(1-p^\star_t)+\pr(X_t\neq\theta) \cdot p^\star_t\nonumber\\
&\geq \pr\p{\mathsf{MAP}_{\calQ}(Z_1^{t-1},X_{t})\neq\theta}\cdot(1-p^\star_t)\nonumber\\
& = \pr_1\p{\mathsf{MAP}_{\calQ}(Z_1^{t-1},X_{t})=2}\cdot(1-p^\star_t),\label{convTotal}
\end{align}
where the last equality follows by symmetry. We next lower bound the probability term at the r.h.s. of \eqref{convTotal}. To this end, we define an event that implies that the output of the mismatched MAP is $2$, given that $\theta=1$. Let $\bar{t}_1,\bar{t}_2\leq t^{\star}$ be three natural numbers, to be defined in the sequel. Let $\mathsf{rev}(t^\star)\triangleq\ppp{i_1,i_2,\ldots,i_{t^\star}}$ be the set of the \emph{last} $t^\star$ revealers. Define the following event 
\begin{align}
\calE(\bar{t}_1,\bar{t}_2,t^{\star})&\triangleq\left\{X_{i}=2,\forall i\in\mathsf{rev}(t^\star)\cup\ppp{i_{\bar{t}_1},i_{\bar{t}_1}+1\ldots,i_{\bar{t}_1}+\bar{t}_2}\right\},
\end{align}
namely, it is the event that the last $t^\star$ revealers are such that their private signal is ``$2$", and all consecutive players $i_{\bar{t}_1},i_{\bar{t}_1}+1,\ldots,i_{\bar{t}_1}+\bar{t}_2$ (either revealers or rationals) are such that their private signal is ``$2$" as well. We claim that by carefully choosing the values of $\bar{t}_1$, $\bar{t}_2$, and $t^\star$, we can show that $\calE(\bar{t}_1,\bar{t}_2,t^{\star})$ implies that the $t$\nth player (mismatched MAP) guess is ``$2$". To this end, note that the first two individuals follow their private signals, that is $Z_1=X_1$ and $Z_2=X_2$. Therefore, if $X_1=X_2=1$, then $\mathsf{R}_2 = (\alpha/\beta)^2$, otherwise, $\mathsf{R}_2 = (\beta/\alpha)^2$. Depending on either one of the above situations, the proceeding guesses depend on whether a student is a revealer or not, and value of the likelihood $\mathsf{R}_{t}$ at each step $t$. Accordingly, at step $i_1-1$, the worst-case (largest) attainable value of the likelihood ratio is
\begin{align}
\mathsf{R}_{i_1-1} &= \p{\frac{\alpha}{\beta}}^2\cdot\prod_{i=3}^{i_1-1}\frac{1-\frac{\beta}{\alpha-\beta}q_i}{1-\frac{\alpha}{\alpha-\beta}q_i}\\&\leq \p{\frac{\alpha}{\beta}}^2e^{\frac{\alpha-\beta}{\alpha+\beta}\sum_{i=3}^{i_1-1}q_i},
\end{align}
which corresponds to the situation where $X_1=X_2=1$, and the proceeding players decisions up to $t\leq i_1-1$ are $Z_t=1$. Now, over $\calE(\bar{t}_1,\bar{t}_2,t^{\star})$, we know that the $i_1$ player is a revealer and its private information is $X_{i_1}=2$. Since he is a revealer its decision will be $Z_{i_1}=2$, which implies that the likelihood ratio is
\begin{align}
\mathsf{R}_{i_1} = \p{\frac{\alpha}{\beta}}\cdot\prod_{i=3}^{i_1-1}\frac{1-\frac{\beta}{\alpha-\beta}q_i}{1-\frac{\alpha}{\alpha-\beta}q_i}.
\end{align}
The index of the next revealer is $i_2$. Thus, since $\mathsf{R}_{i_1}>\alpha/\beta$, the decisions of the proceeding players up to player $i_2$, are ``$1$", which imply that
\begin{align}
\mathsf{R}_{i_2-1} = \p{\frac{\alpha}{\beta}}\cdot\prod_{i=3}^{i_2-1}\frac{1-\frac{\beta}{\alpha-\beta}q_i}{1-\frac{\alpha}{\alpha-\beta}q_i}.
\end{align}
Then, since $i_2$ is a revealer and its private information is $X_{i_2}=2$, we have
\begin{align}
\mathsf{R}_{i_2} = \prod_{i=3}^{i_1-1}\frac{1-\frac{\beta}{\alpha-\beta}q_i}{1-\frac{\alpha}{\alpha-\beta}q_i},
\end{align}
and the above process continues. In particular, assuming that $\mathsf{R}_{i_{j-1}}>\alpha/\beta$, the likelihood ratio after the $i_{j}$\nth decision is
\begin{align}
\mathsf{R}_{i_j} = \p{\frac{\beta}{\alpha}}^{j-2}\prod_{i=3}^{i_j-1}\frac{1-\frac{\beta}{\alpha-\beta}q_i}{1-\frac{\alpha}{\alpha-\beta}q_i}.
\end{align}
We denote by $\bar{t}_1$ the index at which the likelihood ratio $\mathsf{R}_{i_{\bar{t}_1}}$, after the $i_{\bar{t}_1}$\nth decision, is in the interval $\pp{\beta/\alpha,\alpha/\beta}$. Indeed, when this happens, according to \eqref{eqn:mis_like1}--\eqref{eqn:mis_like2}, the likelihood ratio $\mathsf{R}_{i_{\bar{t}_1}}$ will be multiplied by either $\beta/\alpha$ or $\alpha/\beta$ depending on whether the private information is ``$2$" or ``$1$", respectively, until the likelihood ratio value will be either below $\beta/\alpha$ or above $\alpha/\beta$. Accordingly, over $\calE(\bar{t}_1,\bar{t}_2,t^{\star})$, we force the private information of players $i_{\bar{t}_1},i_{\bar{t}_1}+1,\ldots,i_{\bar{t}_1}+\bar{t}_2$ to be ``$2$", so that the corresponding likelihoods will be multiplied by $\beta/\alpha$, until the likelihood ratio value will get bellow $\beta/\alpha$, and accordingly, $\bar{t}_2$ is chosen such that
\begin{align}
\mathsf{R}_{\bar{t}_1+\bar{t}_2} = \p{\frac{\beta}{\alpha}}^{\bar{t}_1+\bar{t}_2-2}\prod_{i=3}^{i_{\bar{t}}-1}\frac{1-\frac{\beta}{\alpha-\beta}q_i}{1-\frac{\alpha}{\alpha-\beta}q_i},
\end{align}
will be less than $\beta/\alpha$. Next, due to the fact that over $\calE(\bar{t}_1,\bar{t}_2,t^{\star})$ the leftover revealers are such that their private information is ``$2$", and the likelihood ratio is below $\beta/\alpha$ (and so MAP outputs ``$2$"), it is clear that all the leftover players decisions will be ``$2$". To assure that we take enough revealers at the end, we chose $t^\star$ such that,
\begin{align}
\p{\frac{\beta}{\alpha}}^{t^\star-2}\prod_{i=3}^{t-t^*}\frac{1-\frac{\beta}{\alpha-\beta}q_i}{1-\frac{\alpha}{\alpha-\beta}q_i}<\frac{\beta}{\alpha},\label{eqn:ell_star_cond}
\end{align}
which reflects the case where the players decision are always ``$1$" up to time $t-t^\star$, and then the left over $t^\star$ players are all revealers with ``$2$" being their private information. It is evident that  \eqref{eqn:ell_star_cond} holds if
\begin{align}
t^\star\geq 3+\frac{\alpha-\beta}{\alpha+\beta}\cdot\frac{\norm{\calQ_t}}{\log(\alpha/\beta)}.
\end{align}
Note that when $\norm{\calQ_t}$ is a finite number which happens to be the case when $q_t=o(t^{-1})$ (as opposed to $\norm{\calP^\star_t}$ which grows logarithmically with $t$), implies that $t^\star$ is finite, which in turn is the reason for the fact that the error probability is finite. Also, by the same token, it is clear that $\bar{t}_1$ and $\bar{t}_2$ are finite too, with $\bar{t}_2<t^\star$. The latter implies that $\calE(\bar{t}_1,t^\star,t^\star)\subseteq\calE(\bar{t}_1,\bar{t}_2,t^\star)$. Finally, we need to make sure that the size of the set of all revealers, denoted by $\mathsf{Rev}_t\triangleq\{i\in[t]:\text{player }i\text{ is revealer}\}$ is bigger than $t^\star$ with high probability. Let $M_t\triangleq\sum_{i=1}^t p_i\sim\log t$. Then, by Chernoff's bound $\pr\p{\abs{\mathsf{Rev}_t}\leq t^\star}\leq\exp(t^\star\log\frac{M_t}{t^\star}-M_t+t^\star) = O(t^{-1})\leq1/2$. Thus, we get that
\begin{align}
\pr_1\p{\mathsf{MAP}_{\calQ}(Z_1^{t-1},X_{t})=2}\geq \pr_1\pp{\calE(\bar{t}_1,\bar{t}_2,t^\star)}
&\geq\sum_{\calA:|\calA|>t^\star}\pr\p{\mathsf{Rev}_t=\calA}\pr_1\pp{\left.\calE(\bar{t}_1,\bar{t}_2,t^\star)\right|\mathsf{Rev}_t=\calA}\nonumber\\
&\geq\sum_{\calA:|\calA|>t^\star}\pr\p{\mathsf{Rev}_t=\calA}\pr_1\pp{\left.\calE(\bar{t}_1,t^\star,t^\star)\right|\mathsf{Rev}_t=\calA}\nonumber\\
& \geq \p{\frac{\beta}{\alpha+\beta}}^{2t^\star}\cdot\pr\p{|\mathsf{Rev}_t|>t^\star}\nonumber\\
&\geq \frac{1}{2}\p{\frac{\beta}{\alpha+\beta}}^{2t^\star}.\label{eqn:lowerBound_cons1}
\end{align}
Therefore, using \eqref{convTotal}, \eqref{eqn:lowerBound_cons1}, and the fact that $\liminf_{t\to\infty}\frac{\norm{\calQ_t}}{\log t}=0$ we get
\begin{align}
\mathsf{E}(\calP^\star,\calQ) &= \liminf_{t\to\infty}-\frac{\log\mathsf{P}_{e,t}(\calP^\star,\calQ)}{\log t}\\
&\leq 2\log\p{\frac{\alpha+\beta}{\beta}}\cdot\liminf_{t\to\infty}\frac{t^\star}{\log t} = 0.
\end{align}
Since it is clear that $\mathsf{E}(\calP^\star,\calQ)\geq0$ we may conclude that in this regime $\mathsf{E}(\calP^\star,\calQ)=0$. Finally, proving that $\mathsf{E}(\calP^\star,\calQ)=0$ for the case $\liminf_{t\to\infty}\frac{\norm{\calQ_t}}{\log t}=\infty$ follows from Theorem~\ref{thm:3} (specifically, using the fact that for $\rho>\rho_1$ we have $\mathsf{E}(\calP^\star,\calQ)=0$), and a monotonicity property of the error probability w.r.t. the revealing probabilities $\calQ$. We provide the complete details in Section~\ref{subsec:lowerBoundRho1}.

\subsection{Proof of Theorem~\ref{thm:3}}\label{subsec:proofTh2}
We split the proofs into several upper and lower bounds, which together characterize tightly the asymptotic learning rate. Note that by ``lower bounds" (``upper bounds") we mean lower- (upper-) bounding the learning rate by upper- (lower-) bounding the error probability. 
\subsubsection{Lower Bound: $\rho_0\leq\rho\leq\rho_1$}\label{subsec:proof_upperBound}
%We emphasize here however, that the inequality,
%\begin{align}
%\inf_{\calP}\mathsf{P}_{e,t}(\calP^\star,\calQ) \geq  \frac{\kappa(\alpha,\beta)}{t},
%\end{align}
%may not be correct due to the following fact: while from the point of view of player $t$ it is clear that MAP is the optimal estimator, it is not clear that in order to minimize his error probability all the other players should employ MAP estimators as well. To wit, it is not clear that 
%\begin{align}
%\inf_{\calQ}\mathsf{P}_{e,t}(\calP,\calQ) = %\mathsf{P}_{e,t}(\calP,\calP),
%\end{align}
%namely, that matched MAP estimation is optimal. 
We analyze next the probability of wrong action by the $t$\nth player. Accordingly, there are two main sources for wrong action: 1) the $t$\nth player is irrational, which happens with probability $p_t$, and his draw is of minority color type, or, 2) the $t$\nth player is rational, but his mismatched MAP estimate is wrong. Accordingly, we can write
\begin{align}
\mathsf{P}_{e,t}(\calP,\calQ) &= \pr\p{\mathsf{MAP}_{\calQ}(Z_1^{t-1},X_{t})\neq\theta}\cdot(1-p_t)+\pr(X_t\neq\theta) \cdot p_t\\
&= \pr\p{\mathsf{MAP}_{\calQ}(Z_1^{t-1},X_{t})\neq\theta}\cdot(1-p_t)+\frac{\beta}{\alpha+\beta} \cdot p_t.\label{eqn:totalExpo}
\end{align}
Therefore, to upper bound $\mathsf{P}_{e,t}(\calP,\calQ)$ we need to upper bound the probability that the mismatched MAP estimator is incorrect. To this end, we next establish some notation, which will simplify the analysis. For $i\in\{1,2\}$ and $t\geq1$, recall our notations in \eqref{notation:1}--\eqref{notation:5}, as well as the definitions in \eqref{eqn:phi_def}--\eqref{eqn:Rtprimedef}. Finally, recall that depending on the value that $\mathsf{R}_{t-1}$ takes, there are three modes of operation for the MAP estimator (see, the paragraph following \eqref{eqn:Rtprimedef}).
In particular, the error probability associated with the MAP estimator can be upper bounded as follows,
\begin{align}
\pr\p{\mathsf{MAP}_{\calQ}(Z_1^{t-1},X_{t})\neq\theta} &= \pr_1\p{\mathsf{MAP}_{\calQ}(Z_1^{t-1},X_{t})\neq1}\nonumber\\
& = \pr_1\p{\mathsf{R}'_t\leq 1}\nonumber\\
&\leq \pr_1\p{\mathsf{R}_{t-1}\leq \frac{\alpha}{\beta}}.\label{eqn:MapProb}
\end{align}
Thus, it is suffice to upper bound $\pr_1\p{\mathsf{R}_{t-1}\leq \alpha/\beta}$. To this end, we use similar techniques used in \cite[Sec. 2.2]{Peres18}, but with modifications which handle the mismatch aspect of our model. Let $\ppp{\calF_t}_{t\geq0}$ denote the filtration spanned by $\ppp{Z_t}_{t\geq1}$. Then, for $\lambda\in[0,1]$, we have
\begin{align}
\bE_1\pp{\left.\p{\frac{\mathsf{R}_t}{\mathsf{R}_{t-1}}}^{-\lambda}\right|\calF_{t-1}}
&=\bE_1\pp{\left.\p{\frac{\hat{\pr}_1(Z_{t}\vert\calF_{t-1})}{\hat{\pr}_2(Z_{t}\vert\calF_{t-1})}}^{-\lambda}\right|\calF_{t-1}}\\
& = \sum_{i\in\{1,2\}}\pr_1(Z_t=i\vert\calF_{t-1})\p{\frac{\hat{\pr}_1(Z_{t}=i\vert\calF_{t-1})}{\hat{\pr}_2(Z_{t}=i\vert\calF_{t-1})}}^{-\lambda},
\end{align}
where $\pr_1(Z_t=i\vert\calF_{t-1})$ is the probability that player $t$ guess is $i$ given the history and $\theta=1$, and the evaluation of this probability is with respect to the underlying true revealers probabilities $\calP$. On the other hand, $\hat{\pr}_1(Z_t=i\vert\calF_{t-1})$ is the probability that player $t$ guess is $i$ given the history and $\theta=1$, and the evaluation of this probability is with respect to the mismatched revealers probabilities $\calQ$. Accordingly, the values of these probabilities are given in \eqref{eqn:mis_like1}--\eqref{eqn:mis_like2}, and
\begin{align}
\pr_1(Z_t=1\vert\calF_{t-1})
&= 
\begin{cases}
\frac{\alpha}{\alpha+\beta}p_t,\ &\;\text{if}\quad\mathsf{R}_{t-1}<\beta/\alpha,\\
\frac{\alpha}{\alpha+\beta},\ &\;\text{if}\quad\mathsf{R}_{t-1}\in[\beta/\alpha,\alpha/\beta],\\
1-\frac{\beta}{\alpha+\beta}p_t,\ &\;\text{if}\quad\mathsf{R}_{t-1}>\alpha/\beta.
\end{cases}\label{eqn:likelihoodP_1}
\end{align}
Therefore, we have,
\begin{align}
\bE_1\pp{\left.\p{\frac{\mathsf{R}_{t}}{\mathsf{R}_{t-1}}}^{-\lambda}\right|\calF_{t-1},\mathsf{R}_{t-1}<\frac{\beta}{\alpha}} &=\frac{\alpha^{1-\lambda}\beta^{\lambda}}{\alpha+\beta}p_t+\p{1-\frac{\alpha}{\alpha+\beta}p_t}\pp{\frac{1-\frac{\beta}{\alpha+\beta}q_t}{1-\frac{\alpha}{\alpha+\beta}q_t}}^{\lambda},\\
\bE_1\pp{\left.\p{\frac{\mathsf{R}_{t}}{\mathsf{R}_{t-1}}}^{-\lambda}\right|\calF_{t-1},\mathsf{R}_{t-1}\in\pp{\frac{\beta}{\alpha},\frac{\alpha}{\beta}}}
&= \frac{\alpha^{1-\lambda}\beta^\lambda+\alpha^\lambda\beta^{1-\lambda}}{\alpha+\beta},\\
\bE_1\pp{\left.\p{\frac{\mathsf{R}_{t}}{\mathsf{R}_{t-1}}}^{-\lambda}\right|\calF_{t-1},\mathsf{R}_{t-1}>\frac{\alpha}{\beta}}
&=\frac{\alpha^\lambda\beta^{1-\lambda}}{\alpha+\beta}p_t+\p{1-\frac{\beta}{\alpha+\beta}p_t}\pp{\frac{1-\frac{\alpha}{\alpha+\beta}q_t}{1-\frac{\beta}{\alpha+\beta}q_t}}^{\lambda}.
\end{align}
Since we assume that both $p_t$ and $q_t$ decay with $t$, we use the fact that $(1-\delta)^\lambda = 1-\lambda\cdot\delta+\Theta(\delta^2)$, as $\delta\to0$. Let
\begin{align}
f_{\lambda}&\equiv f_{\lambda}(\alpha,\beta)\triangleq\frac{\alpha-\alpha^{1-\lambda}\beta^{\lambda}}{\alpha+\beta},\\
g_{\lambda}&\equiv g_{\lambda}(\alpha,\beta)\triangleq \frac{(\alpha-\beta)\lambda}{\alpha+\beta},\\
h_{\lambda}&\equiv h_{\lambda}(\alpha,\beta)\triangleq\frac{\beta-\alpha^{\lambda}\beta^{1-\lambda}}{\alpha+\beta}.
\end{align}
Then, we have
\begin{align}
&\bE_1\pp{\left.\p{\frac{\mathsf{R}_{t}}{\mathsf{R}_{t-1}}}^{-\lambda}\right|\calF_{t-1},\mathsf{R}_{t-1}<\frac{\beta}{\alpha}}= 1-f_{\lambda}\cdot p_t+g_{\lambda}\cdot q_t+O(p_t^2+q_t^2),\label{eqn:taylor1}\\
&\bE_1\pp{\left.\p{\frac{\mathsf{R}_{t}}{\mathsf{R}_{t-1}}}^{-\lambda}\right|\calF_{t-1},\mathsf{R}_{t-1}\in\pp{\frac{\beta}{\alpha},\frac{\alpha}{\beta}}}= \frac{\alpha^{1-\lambda}\beta^\lambda+\alpha^\lambda\beta^{1-\lambda}}{\alpha+\beta},\label{eqn:taylor2}\\
&\bE_1\pp{\left.\p{\frac{\mathsf{R}_{t}}{\mathsf{R}_{t-1}}}^{-\lambda}\right|\calF_{t-1},\mathsf{R}_{t-1}>\frac{\alpha}{\beta}}= 1-h_{\lambda}\cdot p_t-g_{\lambda}\cdot q_t+O(p_t^2+q_t^2).\label{eqn:taylor3}
\end{align}
Define the following two sets:
\begin{align}
\calA_t\triangleq\ppp{i\in[t]:\mathsf{R}_{i-1}>\frac{\alpha}{\beta}},
\end{align}
and $\calB_t\triangleq\calA_t^c=[t]\setminus\calA_{t}$. Define also,
\begin{align}
\mathsf{R}_t^{(1)}\triangleq\prod_{i\in\calA_{t}}\frac{\mathsf{R}_i}{\mathsf{R}_{i-1}}
\end{align}
and
\begin{align}
\mathsf{R}_t^{(2)}\triangleq\prod_{i\in\calB_{t}}\frac{\mathsf{R}_i}{\mathsf{R}_{i-1}}.
\end{align}
Note that $\mathsf{R}_t=\mathsf{R}_t^{(1)}\cdot\mathsf{R}_t^{(2)}$. We these definitions, using \eqref{eqn:taylor1}--\eqref{eqn:taylor3}, we note that there exist some constants $C$ and $C'$ independent of $\lambda$, such that for all $\lambda$,
\begin{align}
\bE_1\pp{\left.\p{\frac{\mathsf{R}_{t}}{\mathsf{R}_{t-1}}}^{-\lambda}e^{h_{\lambda}p_t+g_{\lambda}q_t}\right|\mathsf{R}_{t-1},t\in\calA_t}\leq e^{C'(p_t^2+q_t^2)},
\end{align}
and
\begin{align}
\bE_1\pp{\left.\p{\frac{\mathsf{R}_{t}}{\mathsf{R}_{t-1}}}^{-\lambda}e^{f_{\lambda}p_t-g_{\lambda}q_t}\right|\mathsf{R}_{t-1},t\in\calB_t}\leq e^{C(p_t^2+q_t^2)}.
\end{align}
Recall that $\norm{\calP_t}=\sum_{i=1}^t p_i$ and $\norm{\calQ_t}=\sum_{i=1}^t q_i$. Also, let $\Gamma_t^p\triangleq\sum_{i\in\calA_{t}}^t p_i$ and $\Gamma_t^q\triangleq\sum_{i\in\calA_{t}}^t q_i$. By induction, we can easily see that there exists a constant $C$ such that for any $\lambda_1,\lambda_2\in[0,1]$,
\begin{align}
&\bE_1\left[\p{\mathsf{R}_t^{(1)}}^{-\lambda_1}e^{h_{\lambda_1}\Gamma_t^p+g_{\lambda_1}\Gamma_t^q}\p{\mathsf{R}_t^{(2)}}^{-\lambda_2}\vphantom{\p{\mathsf{R}_t^{(1)}}^{-\lambda_1}}e^{f_{\lambda_2}(\norm{\calP_t}-\Gamma_t^p)-g_{\lambda_2}(\norm{\calQ_t}-\Gamma_t^q)}\right]\leq e^{C\sum_{i=1}^t (p_i^2+q_i^2)},\label{eqn:momentGen}
\end{align}
and since $\sum_{i=1}^t (p_i^2+q_i^2)$ is finite, we can upper bound the r.h.s. of \eqref{eqn:momentGen} by a constant $C_0$. Now, as was shown in \cite[Appendix A]{Peres18}, the condition $\mathsf{R}_t\leq\frac{\alpha}{\beta}$ implies that $\mathsf{R}_t^{(1)}\leq1$. Thus, we may write
\begin{align}
\pr_1\p{\mathsf{R}_t\leq\frac{\alpha}{\beta}}&\leq \pr_1\p{\mathsf{R}_t\leq\frac{\alpha}{\beta},\;t^{-C_1}\leq\mathsf{R}_t^{(1)}\leq1}+\pr_1\p{\mathsf{R}_t^{(1)}\leq t^{-C_1}}.\label{eqn:bound1}
\end{align}
A simple application of multiplicative Chrenoff's bound shows that the second term on the r.h.s. of the above inequality is upper bounded by $t^{-2}$, for some constant $C_1<\infty$. We next upper bound the first term on the r.h.s. of the above inequality. Since $\mathsf{R}_t = \mathsf{R}_t^{(1)}\cdot\mathsf{R}_t^{(2)}$, we can write
\begin{align}
\pr_1\p{\mathsf{R}_t\leq\frac{\alpha}{\beta},\;t^{-C_1}\leq\mathsf{R}_t^{(1)}\leq1}
&\leq\sum_{x=0}^{C_1\log t}\pr_1\p{\mathsf{R}_t^{(1)}\in\pp{e^{-(x+1)},e^{-x}},\mathsf{R}_t^{(2)}\leq e^{x+C_3}}\\
&\leq \sum_{x=0}^{C_1\log t}\pr_1\p{\mathsf{R}_t^{(1)}\leq e^{-x},\mathsf{R}_t^{(2)}\leq e^{x+C_3}},
\end{align}
where $C_3\triangleq 1+\log\frac{\alpha}{\beta}$. Then, for any $\lambda_1,\lambda_2\in[0,1]$, we have
\begin{align}
\pr_1\p{\mathsf{R}_t^{(1)}\leq e^{-x},\mathsf{R}_t^{(2)}\leq e^{x+C_3}}
&= \pr_1\pp{\p{\mathsf{R}_t^{(1)}}^{-\lambda_1}\geq e^{\lambda_1x},\p{\mathsf{R}_t^{(2)}}^{-\lambda_2}\geq e^{-\lambda_2(x+C_3)}},\label{eqn:powerlambda}
\end{align}
and then,
\begin{align}
\pr_1\p{\mathsf{R}_t^{(1)}\leq e^{-x},\mathsf{R}_t^{(2)}\leq e^{x+C_3}}
&= \pr_1\left[\p{\mathsf{R}_t^{(1)}}^{-\lambda_1}e^{h_{\lambda_1}\Gamma_t^p+g_{\lambda_1}\Gamma_t^q}\geq e^{\lambda_1x+h_{\lambda_1}\Gamma_t^p+g_{\lambda_1}\Gamma_t^q},\right.\nonumber\\
&\left.\quad\quad\quad\p{\mathsf{R}_t^{(2)}}^{-\lambda_2}e^{f_{\lambda_2}(\norm{\calP_t}-\Gamma_t^p)-g_{\lambda_2}(\norm{\calQ_t}-\Gamma_t^q)}\right.\nonumber\\
&\left.\quad\quad\quad\geq e^{-(x+C_3)\lambda_2+f_{\lambda_2}(\norm{\calP_t}-\Gamma_t^p)-g_{\lambda_2}(\norm{\calQ_t}-\Gamma_t^q)}\right].\label{eqn:beforeMarkov}
\end{align}
Now using the facts that $\pr[X_1\geq X_2,X_3\geq X_4]\leq\pr[X_1\cdot X_3\geq X_2\cdot X_4]$, for non-negative random variables $X_1^4$, and Markov inequality along with \eqref{eqn:momentGen}, we get
\begin{align}
\pr_1\p{\mathsf{R}_t^{(1)}\leq e^{-x},\mathsf{R}_t^{(2)}\leq e^{x+C_3}}\leq C_0\cdot\bE_1\left[e^{-\lambda_1x-h_{\lambda_1}\Gamma_t^p-g_{\lambda_1}\Gamma_t^q+\lambda_2(x+C_3)}e^{-f_{\lambda_2}(\norm{\calP_t}-\Gamma_t^p)+g_{\lambda_2}(\norm{\calQ_t}-\Gamma_t^q)}\right].\label{eqn:Markov}
\end{align}
Using the facts that $\norm{\calQ_t} = \rho\cdot\norm{\calP_t}$ and $\Gamma_t^q = \rho\cdot\Gamma_t^p$, we obtain,
\begin{align}
\pr_1\p{\mathsf{R}_t^{(1)}\leq e^{-x},\mathsf{R}_t^{(2)}\leq e^{x+C_3}} C_0\cdot\bE_1\left[e^{x(\lambda_2-\lambda_1)+\lambda_2C_3-(f_{\lambda_2}-\rho g_{\lambda_2})\norm{\calP_t}} e^{(f_{\lambda_2}-\rho g_{\lambda_2}-h_{\lambda_1}-\rho g_{\lambda_1})\Gamma_t^p}\right].
\end{align}
By symmetry, we can get the above upper bound with $(\lambda_2-\lambda_1)$ replaced by $(\lambda_1-\lambda_2)$. Indeed, to show this we replace \eqref{eqn:powerlambda} with
\begin{align}
\pr_1\p{\mathsf{R}_t^{(1)}\leq e^{-x},\mathsf{R}_t^{(2)}\leq e^{x+C_3}}
= \pr_1\left[\p{\mathsf{R}_t^{(1)}}^{-(1-\lambda_1)}\geq e^{(1-\lambda_1)x},\p{\mathsf{R}_t^{(2)}}^{-(1-\lambda_2)}\geq e^{-(1-\lambda_2)(x+C_3)}\right],\label{eqn:powerlambda2}
\end{align}
and follow the \eqref{eqn:beforeMarkov}--\eqref{eqn:Markov}. Therefore, we may write
\begin{align}
\pr_1\p{\mathsf{R}_t^{(1)}\leq e^{-x},\mathsf{R}_t^{(2)}\leq e^{x+C_3}}\leq C_0\cdot\bE_1\left[e^{-x\abs{\lambda_2-\lambda_1}+\lambda_2C_3-(f_{\lambda_2}-\rho g_{\lambda_2})\norm{\calP_t}} e^{(f_{\lambda_2}-\rho g_{\lambda_2}-h_{\lambda_1}-\rho g_{\lambda_1})\Gamma_t^p}\right].
\end{align}
We can now optimize our choices of $\lambda_1$ and $\lambda_2$ to minimize the above upper bound. We take $\lambda_1+\lambda_2=1$. For such a pair it is easy to check that,
\begin{align}
f_{\lambda_2}-\rho g_{\lambda_2}-h_{\lambda_1}-\rho g_{\lambda_1} = \frac{\alpha-\beta}{\alpha+\beta}(1-\rho).\label{eqn:rhsBrackers}
\end{align}
We next consider the case where $\rho\geq1$, for which the r.h.s. of \eqref{eqn:rhsBrackers} is negative, and so,
\begin{align}
\pr_1\p{\mathsf{R}_t^{(1)}\leq e^{-x},\mathsf{R}_t^{(2)}\leq e^{x+C_3}}
& \leq C_0\cdot\bE_1\left[e^{-x|2\lambda_1-1|+\lambda_2C_3-(f_{1-\lambda_1}-\rho g_{1-\lambda_1})\norm{\calP_t}} e^{\frac{\alpha-\beta}{\alpha+\beta}(1-\rho)\Gamma_t^p}\right]\\
&\leq C_0\cdot e^{-x|2\lambda_1-1|+(1-\lambda_1)C_3-(f_{1-\lambda_1}-\rho g_{1-\lambda_1})\norm{\calP_t}}.
\end{align}
In the interval $\lambda_1\in[0,1]$, it can be checked that $f_{1-\lambda_1}-\rho g_{1-\lambda_1}$ is maximized at 
\begin{align}
\lambda_1^\star = \min\p{1,\frac{\log\p{\rho\cdot\frac{\alpha/\beta-1}{\log\alpha/\beta}}}{\log\alpha/\beta}}.\label{eqn:lambdastarDef}
\end{align}
We mention here that $\lambda_1^\star$ satisfies the following equality
\begin{align}
\p{\frac{\alpha}{\beta}}^{\lambda_1^\star} = \frac{(\alpha-\beta)\rho}{\beta\log\frac{\alpha}{\beta}},
\end{align}
which proves to be useful. It can be shown that $\lambda_1^\star\geq1/2$. Thus, whenever $\lambda_1^\star<1$, which happens to be the case exactly when $\rho\leq \rho_{1}$, we have
\begin{align}
\pr_1\p{\mathsf{R}_t^{(1)}\leq e^{-x},\mathsf{R}_t^{(2)}\leq e^{x+C_3}}
&\leq C_0\cdot e^{-x|2\lambda_1^\star-1|+(1-\lambda_1^\star)C_3-(f_{1-\lambda_1^\star}-\rho g_{1-\lambda_1^\star})\norm{\calP_t}}.
\end{align}
It can be checked that
\begin{align}
f_{1-\lambda_1^\star}-\rho g_{1-\lambda_1^\star}&=\frac{\alpha\log\frac{\alpha}{\beta}-\rho(\alpha-\beta)}{(\alpha+\beta)\log\frac{\alpha}{\beta}}-\rho\frac{\alpha-\beta}{\alpha+\beta}(1-\lambda_1^\star)\\
& =\frac{\alpha\log\frac{\alpha}{\beta}-\rho(\alpha-\beta)}{(\alpha+\beta)\log\frac{\alpha}{\beta}}-\frac{\rho(\alpha-\beta)}{\alpha+\beta}\pp{1-\frac{\log\pp{\frac{\rho(\alpha/\beta-1)}{\log\frac{\alpha}{\beta}}}}{\log\frac{\alpha}{\beta}}}\\
& = \frac{\frac{\alpha}{\beta}\log\frac{\alpha}{\beta}-\rho(\frac{\alpha}{\beta}-1)\pp{1+\log\frac{\frac{\alpha}{\beta}\log\frac{\alpha}{\beta}}{\rho(\frac{\alpha}{\beta}-1)}}}{(1+\frac{\alpha}{\beta})\log\frac{\alpha}{\beta}}\\
&=\delta(\alpha/\beta,\rho),
\end{align}
where $\delta(\alpha/\beta,\rho)$ is defined \eqref{eqn:delta_def}. Combining the above result with \eqref{eqn:bound1}, we get
\begin{align}
\pr_1\p{\mathsf{R}_t\leq\frac{\alpha}{\beta}}\leq C_0'e^{(1-\lambda_1^\star)C_3-\delta(\alpha/\beta,\rho)\norm{\calP_t}}+\frac{1}{t^2},
\end{align}
where we have used the fact that $\sum_{x=0}^{C_1\log t}e^{-x|2\lambda_1^\star-1|}$ is finite, and absorbed its value in the constant $C_0'$. Then, substituting the above result in \eqref{eqn:MapProb} and then in \eqref{eqn:totalExpo}, we obtain
\begin{align}
\mathsf{P}_{e,t}(\calP,\calQ) &\leq \pp{C_0'e^{(1-\lambda_1^\star)C_3-\delta(\gamma,\rho)\norm{\calP_t}}+\frac{1}{t^2}}\cdot(1-p_t)+\frac{\beta}{\alpha+\beta} \cdot p_t.\label{eqn:upperBounderrorgen1}
\end{align}
Therefore, taking $p_t = \frac{(1+\gamma)\kappa(\gamma)}{t}\wedge 1 = p^\star_t$, we obtain that for $1\leq \rho\leq \rho_{1}$,
\begin{align}
\mathsf{E}(\calP^\star,\calQ)&=\liminf_{t\to\infty}-\frac{\log\mathsf{P}_{e,t}(\calP^\star,\calQ)}{\log t}\\
&\geq \delta(\alpha/\beta,\rho)\cdot\liminf_{t\to\infty}\frac{\norm{\calP_t}}{\log t} \\
&= \delta(\gamma,\rho)\pp{(1+\gamma)\kappa(\gamma)},
\end{align}
as claimed. Next, we consider the case where $\rho<1$. In this case, the the r.h.s. of \eqref{eqn:rhsBrackers} is positive, and so,
\begin{align}
\pr_1\p{\mathsf{R}_t^{(1)}\leq e^{-x},\mathsf{R}_t^{(2)}\leq e^{x+C_3}}
&\leq C_0\cdot\bE_1\left[e^{-x|2\lambda_1-1|+\lambda_2C_3-(f_{1-\lambda_1}-\rho g_{1-\lambda_1})\norm{\calP_t}} e^{\frac{\alpha-\beta}{\alpha+\beta}(1-\rho)\Gamma_t^p}\right]\\
&\leq C_0\cdot e^{-x|2\lambda_1-1|+(1-\lambda_1)C_3-(f_{1-\lambda_1}-\rho g_{1-\lambda_1})\norm{\calP_t}}\cdot e^{\frac{\alpha-\beta}{\alpha+\beta}(1-\rho)\norm{\calP_t}}.
\end{align}
Again, $f_{1-\lambda_1}-\rho g_{1-\lambda_1}$ is maximized at 
\begin{align}
\lambda_1^\star = \frac{\log\p{\rho \cdot\frac{\alpha/\beta-1}{\log\alpha/\beta}}}{\log\alpha/\beta},
\end{align}
and note that for $\rho<1$ it is always the case that $\lambda_1^\star<1$. Also, for $\rho_{0}<\rho\leq 1$, we have that $\lambda_1^\star\in[0,1]$. Hence, we may write
\begin{align}
\pr_1\p{\mathsf{R}_t^{(1)}\leq e^{-x},\mathsf{R}_t^{(2)}\leq e^{x+C_3}}& \leq C_0e^{-x|2\lambda_1^\star-1|+(1-\lambda_1^\star)C_3-\pp{\delta(\alpha/\beta,\rho)-\frac{\alpha-\beta}{\alpha+\beta}(1-\rho)}\norm{\calP_t}}.
\end{align}
Combining the above result with \eqref{eqn:bound1}, we get
\begin{align}
\pr_1\p{\mathsf{R}_t\leq\frac{\alpha}{\beta}}&\leq C_0'e^{(1-\lambda_1^\star)C_3-\pp{\delta(\gamma,\rho)-\frac{\alpha-\beta}{\alpha+\beta}(1-\rho)}\norm{\calP_t}}+\frac{1}{t^2}.
\end{align}
Then, substituting the above result in \eqref{eqn:MapProb} and then in \eqref{eqn:totalExpo}, we obtain
\begin{align}
\mathsf{P}_{e,t}(\calP,\calQ) \leq \frac{\beta}{\alpha+\beta} \cdot p_t+\pp{C_0'e^{(1-\lambda_1^\star)C_3-\pp{\delta(\gamma,\rho)-\frac{\alpha-\beta}{\alpha+\beta}(1-\rho)}\norm{\calP_t}}+\frac{1}{t^2}}\cdot(1-p_t).\label{eqn:upperBounderrorgen2}
\end{align}
Therefore, taking $p_t = \frac{\alpha+\beta}{\beta}\frac{\kappa(\alpha,\beta)}{t}\wedge 1 = p^\star_t$, we obtain that for $\rho_{0}\leq \rho\leq 1$,
\begin{align}
\mathsf{E}(\calP^\star,\calQ)&=\liminf_{t\to\infty}-\frac{\log\mathsf{P}_{e,t}(\calP^\star,\calQ)}{\log t}\nonumber\\
&\geq \pp{\delta(\gamma,\rho)-(1-\rho)\frac{\gamma-1}{\gamma+1}}\cdot\liminf_{t\to\infty}\frac{\norm{\calP_t^\star}}{\log t} \nonumber\\
&= \pp{\delta(\gamma,\rho)-(1-\rho)\frac{\gamma-1}{\gamma+1}}(1+\gamma)\kappa(\gamma),
\end{align} 
as claimed.

\subsubsection{Upper Bound: $\rho\geq\rho_1$ and $\norm{\calQ_t}\gg\log t$}\label{subsec:lowerBoundRho1}
We prove that for $\rho\geq\rho_1$ we have $\mathsf{E}(\calP^\star,\calQ)=0$. We show that this is correct also when $\liminf_{t\to\infty}\frac{\norm{\calQ_t}}{\log t}=\infty$ as stated in Theorem~\ref{thm:2}. To this end, first note that from \eqref{notation:5}, we have
\begin{align}
\pr\p{\mathsf{MAP}_{\calQ}(Z_1^{t-1},X_{t})\neq\theta} &= \pr_1\p{\mathsf{MAP}_{\calQ}(Z_1^{t-1},X_{t})\neq1}\nonumber\\
& \geq \pr_1\p{\mathsf{R}'_t< 1}\nonumber\\
&\geq \pr_1\p{\mathsf{R}_{t-1}< \frac{\beta}{\alpha}},\label{eqn:MapProb2}
\end{align}
and so it is suffice to lower bound the r.h.s. of \eqref{eqn:MapProb2}. It is clear that 
\begin{align}
\pr_1\p{\mathsf{R}_{t}< \frac{\beta}{\alpha}}\geq \pr_1\p{\mathsf{R}_i<\frac{\beta}{\alpha},\;\forall i\in[t]}.\label{eqn:lowerBoundIntuition}
\end{align}
Accordingly, in order to obtain a lower bound on \eqref{eqn:lowerBoundIntuition} we define three events that together imply that $\mathsf{R}_{t}<\beta/\alpha$. We note that the derivations bellow follow \cite[Sec. 2.2]{Peres18}, with modifications which handle the mismatch aspect of our model. We need a few definitions. Let $\tau(s)\triangleq\min\{t\geq1:\norm{\calQ_t}\geq s\}$, and $t_0\triangleq\tau(2\frac{\alpha-\beta}{\alpha+\beta}\log\frac{\alpha}{\beta}+2)$. Define 
\begin{align}
\calE_0\triangleq\ppp{\mathsf{R}_{t_0}<\p{\beta/\alpha}^4}.
\end{align}
The above initial event takes the mismatched likelihood ratio below $\beta/\alpha$, and the events we define below ensure that it always stays below this bar. Let $\mathsf{J}_t\triangleq\log\mathsf{R}_t$, and define the stopping time $T\triangleq\min\{s\geq t_0:J_s\not\in[-\log t,2\log(\beta/\alpha)]\}$. We define the events
\begin{align}
\calE_1\triangleq\ppp{\mathsf{J}_T\leq-\log t},
\end{align}
and
\begin{align}
\calE_2\triangleq\ppp{\min_{s\in[t]}\mathsf{J}_s\geq-\log^{3/4} t}.
\end{align}
We observe that $\calE_0\cap\calE_1\cap\calE_2$ imply together that $\mathsf{J}_s\in\pp{-\log^{3/4} t,2\log\frac{\beta}{\alpha}}$, for all $s\in[t_0,t]$, which in turn implies that $\mathsf{R}_t<\beta/\alpha$. Thus,
\begin{align}
\pr_1\p{\mathsf{R}_{t}\leq \frac{\beta}{\alpha}}\geq\pr_1\p{\calE_0\cap\calE_1\cap\calE_2}.\label{eqn:lowerboundMismatchByE0E1E2}
\end{align}
We next lower bound the probability of the event $\calE_0$ which is easier to handle. Note that according to our setting the first two individuals follow their private signal, and hence if $X_1=X_2=2$, we have $Z_1=Z_2=2$. This in turn implies that $\mathsf{R}_2 = (\beta/\alpha)^2$. Now, if $X_i=2$ for all $i\in\{3,4,\ldots,t_0\}$, then it is clear that $Z_i=2$, for all $i\in\{3,4,\ldots,t_0\}$ as well. Accordingly, using \eqref{eqn:mis_like1}--\eqref{eqn:mis_like2}, this implies that the mismatched likelihood ratio at time $t_0$ is given by
\begin{align}
\mathsf{R}_{t_0} &= \p{\frac{\beta}{\alpha}}^2\prod_{i=3}^{t_0}\frac{1-\frac{\alpha}{\alpha+\beta}q_i}{1-\frac{\beta}{\alpha+\beta}q_i}\\
&\leq \p{\frac{\beta}{\alpha}}^2\exp\p{-\frac{\alpha-\beta}{\alpha+\beta}\sum_{i=3}^{t_0}q_i}.\label{eqn:upperBoundLikelihoodt_0}
\end{align}
However, by the definition of $t_0$, we know that
\begin{align}
\sum_{i=3}^{t_0}q_i\geq \norm{\calQ_{t_0}}-2\geq 2\frac{\alpha+\beta}{\alpha-\beta}\log\frac{\alpha}{\beta},
\end{align}
which together with \eqref{eqn:upperBoundLikelihoodt_0} implies that $\mathsf{R}_{t_0}\leq(\beta/\alpha)^4$. Thus,
\begin{align}
\pr_1\p{\calE_0}\geq \pr_1\p{X_i=2\;\forall i\in[t_0]} = \p{\frac{\beta}{\alpha+\beta}}^{t_0}.\label{eqn:lowerboundE0}
\end{align}
Therefore, because $t_0$ is a constant it is suffice to lower bound the probability $\pr_1(\calE_1\cap\calE_2\vert\calE_0)$. Given $\calE_0$, the log-likelihood ratio $\mathsf{J}_t$ performs a random walk from time $t_0$ until the stopping time $T$. Specifically, using \eqref{eqn:mis_like1}--\eqref{eqn:mis_like2}, and  \eqref{eqn:likelihoodP_1}, for $s\geq t_0$, we may write
\begin{align}
\mathsf{J}_{s\wedge T} = \mathsf{J}_{t_0}+\sum_{i=t_0+1}^{s\wedge T}\xi_i
\end{align}
where $\ppp{\xi_i}$ are statistically independent random variables such that
\begin{align}
\pr_1\p{\xi_i=\log\frac{\alpha}{\beta}} = \frac{\alpha}{\alpha+\beta}p_i,\label{eqn:randomwalk1}
\end{align}
and
\begin{align}
\pr_1\p{\xi_i=\log\frac{1-\frac{\alpha}{\alpha+\beta}q_i}{1-\frac{\beta}{\alpha+\beta}q_i}} = 1-\frac{\alpha}{\alpha+\beta}p_i.\label{eqn:randomwalk2}
\end{align}
Now, note that
\begin{align}
\bE_1\pp{\xi_i} &= \frac{\alpha}{\alpha+\beta}\log\frac{\alpha}{\beta}p_i-\frac{\alpha-\beta}{\alpha+\beta}q_i+\Theta(q_i^2+p_i^2)\nonumber\\
& = \frac{\rho+\gamma(\log\gamma-\rho)}{1+\gamma}p_i+\Theta(p_i^2)\label{eqn:downwardRandomWalk}
\end{align}
where $\gamma=\alpha/\beta$. The important observation here is that $\rho\geq\rho_1$ is equivalent $\rho+\gamma(\log\gamma-\rho)\leq0$, which implies that the log-likelihood ratio has a downward (non-positive) drift. More precisely, it can be seen that the expectation can be written as $\bE_1[\xi_i] = -\eta\cdot p_i+\Theta(p_i^2)$, for some $\eta>0$. Since $p_i$ is decaying with $i$, it is clear that there exists a \emph{finite} index $i_0\in\mathbb{N}$, such that $\bE_1[\xi_i]\leq0$, for all $i\geq i_0$. Accordingly, letting $\bar{t}_0\triangleq t_0\vee i_0$, we obtain that under $\pr_1$, the random walk $\ppp{\mathsf{J}_{s\wedge T}}_{s\geq \bar{t}_0}$ is a supermartingale. For simplicity of notation, for the rest of the proof we use $t_0$ in place of $\bar{t}_0$. Therefore, by the optional stopping theorem we have 
\begin{align}
\bE_1[\mathsf{J}_T\vert\calE_0]\leq \bE_1[\mathsf{J}_{t_0}]\leq 4\log\frac{\beta}{\alpha}.\label{eqn:epxectationUpperBound}
\end{align}
On the other hand, by the definition of $T$, it is either the case that $\mathsf{J}_T>2\log\frac{\beta}{\alpha}$, in which case $\mathsf{J}_T\in(2\log\frac{\beta}{\alpha},\log\frac{\beta}{\alpha}]$, or $\mathsf{J}_T<-\log t$, and then $\mathsf{J}_T\in[-\log t-\log\frac{\beta}{\alpha},-\log t)$. Thus, we can write
\begin{align}
\bE_1[\mathsf{J}_T\vert\calE_0] &=\bE_1\pp{\mathsf{J}_T\mathds{1}\pp{\mathsf{J}_T>2\log\frac{\beta}{\alpha}}\vert\calE_0}+\bE_1\pp{\mathsf{J}_T\mathds{1}\pp{\mathsf{J}_T<-\log t}\vert\calE_0} \\
&\geq \pp{1-\pr_1(\calE_1\vert\calE_0)}2\log\frac{\beta}{\alpha}-\pr_1(\calE_1\vert\calE_0)\cdot\p{\log t+\log\frac{\beta}{\alpha}}\\
&\geq 2\log\frac{\beta}{\alpha}-\pr_1(\calE_1\vert\calE_0)\cdot\log t,\label{eqn:epxectationLowerBound}
\end{align}
which together with \eqref{eqn:epxectationUpperBound} implies that
\begin{align}
\pr_1(\calE_1\vert\calE_0)\geq\frac{2\log\gamma}{\log t}.
\end{align}
Finally, using classical results on the tails of supermartingales (see, e.g., \cite{freedman1975,fan2015}), we have
\begin{align}
\pr\p{\left.\min_{s\in[t]}\mathsf{J}_s<-\log^{3/4}t\right|\calE_0}\leq e^{-c\sqrt{\log t}},
\end{align}
and thus
\begin{align}
\pr_1\p{\calE_1\cap\calE_2\vert\calE_0}\geq \frac{\log\gamma}{\log t},\label{eqn:lowerboundE1E2}
\end{align}
for $t$ large enough. Combining \eqref{convTotal}, \eqref{eqn:MapProb2}, \eqref{eqn:lowerboundMismatchByE0E1E2}, \eqref{eqn:lowerboundE0}, and \eqref{eqn:lowerboundE1E2}, we obtain
\begin{align}
\mathsf{E}(\calP^\star,\calQ) &= \liminf_{t\to\infty}-\frac{\log\mathsf{P}_{e,t}(\calP^\star,\calQ)}{\log t}\\
&\leq \liminf_{t\to\infty}-\frac{\log\pr_1\p{\calE_1\cap\calE_2\vert\calE_0}}{\log t} = 0,
\end{align}
which concludes the proof. Finally, using the above arguments we prove that when $\bar\calQ$ satisfies $\liminf_{t\to\infty}\frac{\norm{\bar\calQ_t}}{\log t}=\infty$, then $\mathsf{E}(\calP^\star,\bar\calQ)=0$, as stated in Theorem~\ref{thm:2}. Specifically, let $\calQ$ be any sequence of assumed revealing probabilities such that ${q}_t = \rho\cdot p_t^\star$, with $\rho>\rho_1$. Let $\mathsf{R}^{\bar\calQ}_t$ and $\mathsf{R}^{\calQ}_t$ designate the likelihoods corresponding to the revealing probabilities $\bar\calQ$ and $\calQ$, respectively. Then, from \eqref{eqn:lowerBoundIntuition} it is clear that
\begin{align}
\pr\p{\mathsf{MAP}_{\bar\calQ}(Z_1^{t},X_{t+1})\neq\theta}
&\geq \pr_1\p{\mathsf{R}^{\bar\calQ}_{t}< \frac{\beta}{\alpha}}\\
&\geq \pr_1\p{\mathsf{R}^{\bar\calQ}_i<\frac{\beta}{\alpha},\;\forall i\in[t]}.
\end{align}
Now, note that for large enough $t$ it must be the case that $\bar{q}_t>q_t$. The log-likelihood ratio $\mathsf{J}_t^{\bar{\calQ}}$ performs a random walk with probabilities given in \eqref{eqn:randomwalk1}--\eqref{eqn:randomwalk2}, with $q_t$ replaced by $\bar{q}_t$. Accordingly, it is clear that the random variables $\{\xi_i\}$ can take only smaller values under $\bar\calQ$ compared to $\calQ$. This in turn implies that $\mathsf{R}^{\bar\calQ}\leq\mathsf{R}^{\calQ}$, and thus,
\begin{align}
\pr\p{\mathsf{MAP}_{\bar\calQ}(Z_1^{t},X_{t+1})\neq\theta}
&\geq \pr_1\p{\mathsf{R}^{\bar\calQ}_i<\frac{\beta}{\alpha},\;\forall i\in[t]}\nonumber\\
&\geq \pr_1\p{\mathsf{R}^{\calQ}_i<\frac{\beta}{\alpha},\;\forall i\in[t]}\nonumber\\
&\geq \pr_1(\calE_0\cap\calE_1\cap\calE_2),
\end{align}
which is the same lower bound we started with for $\calQ$.

\subsubsection{Upper Bound: $1\leq\rho\leq\rho_1$}\label{subsec:lowerBoundRho1_3}
We consider the case where $1\leq\rho\leq\rho_1$, and continue from \eqref{eqn:downwardRandomWalk}. Indeed, in this regime, the expectation in \eqref{eqn:downwardRandomWalk} is non-negative and thus the log-likelihood ratio $\mathsf{J}_s$ has an upward drift. We remove this drift by defining a new measure $\tilde{\pr}_1$, such that for $i>t_0$,
\begin{align}
\tilde{\pr}_1\p{\xi_i=\log\frac{\alpha}{\beta}} &= \nu_i,\\
\tilde{\pr}_1\p{\xi_i=\log\frac{1-\frac{\alpha}{\alpha+\beta}q_i}{1-\frac{\beta}{\alpha+\beta}q_i}} &= 1-\nu_i,
\end{align}
where 
\begin{align}
\nu_i\triangleq\frac{\log\frac{1-\frac{\beta}{\alpha+\beta}q_i}{1-\frac{\alpha}{\alpha+\beta}q_i}}{\log\p{\frac{\alpha}{\beta}\frac{1-\frac{\beta}{\alpha+\beta}q_i}{1-\frac{\alpha}{\alpha+\beta}q_i}}}.\label{eqn:nu_idef}
\end{align}
Now, under $\tilde{\pr}_1$, the random walk $\ppp{\mathsf{J}_{s\wedge T}}_{s\geq t_0}$ is a martingale, and thus using the same steps we used in \eqref{eqn:epxectationUpperBound}--\eqref{eqn:lowerboundE1E2}, we obtain that
\begin{align}
\tilde{\pr}_1\p{\calE_1\cap\calE_2\vert\calE_0}\geq \frac{\log\gamma}{\log t},
\end{align}
for $t$ large enough. Next, performing a change of measure we may write
\begin{align}
\pr_1\p{\calE_1\cap\calE_2\vert\calE_0} = \tilde{\bE}_1\pp{\left.\frac{\mathrm{d}\pr_1(\cdot\vert\calE_0)}{\mathrm{d}\tilde{\pr}_1(\cdot\vert\calE_0)}\mathds{1}\pp{\calE_1\cap\calE_2}\right|\calE_0},\label{eqn:ChangeOfMeasure}
\end{align}
so we need to understand how the Radon-Nikodym derivative of $\pr_1(\cdot\vert\calE_0)$ w.r.t. $\tilde{\pr}_1(\cdot\vert\calE_0)$ behaves. Note that
\begin{align}
&\frac{\mathrm{d}\pr_1(\cdot\vert\calE_0)}{\mathrm{d}\tilde{\pr}_1(\cdot\vert\calE_0)} = \prod_{i=t_0+1}^t\left\{\frac{\frac{\alpha}{\alpha+\beta}p_i^\star}{\nu_i}\mathds{1}\pp{\xi_i=\log\frac{\alpha}{\beta}}+\frac{1-\frac{\alpha}{\alpha+\beta}p_i^\star}{1-\nu_i}\mathds{1}\pp{\xi_i=\log\frac{1-\frac{\alpha}{\alpha+\beta}q_i}{1-\frac{\beta}{\alpha+\beta}q_i}}\right\}.
\end{align}
We claim that each factor in the product can be lower bounded for some $C=C(\alpha,\beta)$ as follows
\begin{align}
&\frac{\frac{\alpha}{\alpha+\beta}p_i^\star}{\nu_i}\mathds{1}\pp{\xi_i=\log\frac{\alpha}{\beta}}+\frac{1-\frac{\alpha}{\alpha+\beta}p_i^\star}{1-\nu_i}\mathds{1}\pp{\xi_i=\log\frac{1-\frac{\alpha}{\alpha+\beta}q_i}{1-\frac{\beta}{\alpha+\beta}q_i}}\geq e^{(1-\lambda^\star)\xi_i}\cdot K_i(\xi_i)\label{eqn:RadonInequality}
\end{align}
where $\lambda^\star=\lambda_1^\star$ is defined in \eqref{eqn:lambdastarDef}, and
\begin{align}
K_i(\xi_i)&\triangleq e^{-\delta(\gamma,\rho)p_i^\star-C(p_i^\star)^2}\cdot\mathds{1}\pp{\xi_i=\log\frac{1-\frac{\alpha}{\alpha+\beta}q_i}{1-\frac{\beta}{\alpha+\beta}q_i}}+e^{-\rho\p{\frac{1}{2}-\frac{\alpha-\beta}{(\alpha+\beta)\log\frac{\alpha}{\beta}}}p_i^\star-C(p_i^\star)^2}\cdot\mathds{1}\pp{\xi_i=\log\frac{\alpha}{\beta}}.
\end{align}
Indeed, this inequality can be checked for both potential values of $\xi$ by expanding the expressions in $p_i^\star$. Then, multiplying \eqref{eqn:RadonInequality} over all $i\in\{t_0+1,\ldots,t\}$, using the fact that $\mathsf{J}_t = \mathsf{J}_{t_0}+\sum_{i=t_0+1}^t\xi_i$ on the event $\calE_1\cap\calE_2$, we obtain that
\begin{align}
\frac{\mathrm{d}\pr_1(\cdot\vert\calE_0)}{\mathrm{d}\tilde{\pr}_1(\cdot\vert\calE_0)}\mathds{1}\pp{\calE_1\cap\calE_2}&\geq e^{(1-\lambda^\star)(\mathsf{J}_{t}-\mathsf{J}_{t_0})-C\sum_{i=1}^t(p_i^\star)^2}\cdot e^{-\delta(\gamma,\rho)\sum_{i\in\calV^c}p_i^\star-\rho\p{\frac{1}{2}-\frac{\alpha-\beta}{(\alpha+\beta)\log\frac{\alpha}{\beta}}}\sum_{i\in\calV}p_i^\star}\mathds{1}\pp{\calE_1\cap\calE_2}\nonumber\\
&\geq e^{(1-\lambda^\star)\mathsf{J}_{t}-C'}\cdot e^{-\delta(\gamma,\rho)\norm{\calP_t^\star}-\rho\p{\frac{1}{2}-\frac{\alpha-\beta}{(\alpha+\beta)\log\frac{\alpha}{\beta}}}\sum_{i\in\calV}p_i^\star}\mathds{1}\pp{\calE_1\cap\calE_2},
\end{align}
where $\calV\triangleq\ppp{i\geq t_0:\xi_i=\log(\alpha/\beta)}$, $C'\triangleq C\sum_{i=1}^t(p_i^\star)^2$ is finite, and the second inequality follows because conditioned on $\calE_0$ we know that $\mathsf{J}_{t_0}<0$. Also, recall that on the event $\calE_1\cap\calE_2$ we have that $\mathsf{J}_t\geq-\log^{3/4}t$, and thus
\begin{align}
&\frac{\mathrm{d}\pr_1(\cdot\vert\calE_0)}{\mathrm{d}\tilde{\pr}_1(\cdot\vert\calE_0)}\mathds{1}\pp{\calE_1\cap\calE_2}\geq e^{-(1-\lambda^\star)\log^{3/4}t-C'-\delta(\gamma,\rho)\norm{\calP_t^\star}} e^{-\rho\p{\frac{1}{2}-\frac{\alpha-\beta}{(\alpha+\beta)\log\frac{\alpha}{\beta}}}\sum_{i\in\calV}p_i^\star}\mathds{1}\pp{\calE_1\cap\calE_2}.\label{eqn:Radon_lower_2}
\end{align}
We next show that with high probability $\abs{\calV}$ is at most logarithmic in $t$ and thus $\sum_{i\in\calV}p_i^\star$ is negligible compared to other contributions in the exponent of the r.h.s. of \eqref{eqn:Radon_lower_2}. Let $Z\triangleq \sum_{i=t_0+1}^t\mathds{1}\pp{\xi_i=\log\frac{\alpha}{\beta}}$. Then, we already saw that under $\tilde{\pr}_1$ the random variables $\ppp{\xi}_{i>t_0}$ are statistically independent. Specifically, $\mathsf{Z}$ follows a Poisson-Binomial distribution with success probabilities $\nu_i$ given in \eqref{eqn:nu_idef}. Using Chernoff's inequality, for a Poisson-Binomial random variable $\mathsf{Z}$ with mean $\mu$, and any $s>\mu$, it can be shown that
\begin{align}
\pr\pp{\mathsf{Z}\geq s}\leq \exp\p{s-\mu-s\log\frac{s}{\mu}}.\label{eqn:poissbinomTails}
\end{align}
Accordingly, in our case it is clear that $\mu = \sum_{i=t_0+1}^t\nu_i = C_1(1+o(1))\cdot\log t$, as $t\to\infty$, for some $C_1(\alpha,\beta)$, due to the fact that $\nu_i\propto q_i = \Theta(t^{-1})$. Taking $s=\ell\cdot\mu$, such that $(\ell-1-\ell\log\ell)\leq -\frac{2}{C_1}$, we obtain from \eqref{eqn:poissbinomTails} that $\tilde{\pr}_1\pp{\mathsf{Z}\geq s}\leq t^{-2}$. Thus, with probability at least $1-O(t^{-2})$ we have that $\abs{\calV}\leq C''\log t$, for some constant $C''$. This in turn implies that with the same probability
\begin{align}
\sum_{i\in\calV}p_i^\star\leq C_2(1+o(1))\cdot\log(\log t) = o(\log t).\label{eqn:loglogBehaviour}
\end{align}
Therefore, combining \eqref{eqn:ChangeOfMeasure}, \eqref{eqn:Radon_lower_2}, and \eqref{eqn:loglogBehaviour}, we obtain
\begin{align}
\pr_1\p{\calE_1\cap\calE_2\vert\calE_0}&= \tilde{\bE}_1\pp{\left.\frac{\mathrm{d}\pr_1(\cdot\vert\calE_0)}{\mathrm{d}\tilde{\pr}_1(\cdot\vert\calE_0)}\mathds{1}\pp{\calE_1\cap\calE_2}\right|\calE_0}\nonumber\\
&\geq \tilde{\bE}_1\left[\left.e^{-(1-\lambda^\star)\log^{3/4}t-C'-\delta(\gamma,\rho)\norm{\calP_t^\star}}\cdot e^{-\rho\p{\frac{1}{2}-\frac{\alpha-\beta}{(\alpha+\beta)\log\frac{\alpha}{\beta}}}\sum_{i\in\calV}p_i^\star}\mathds{1}\pp{\calE_1\cap\calE_2}\right|\calE_0\right]\nonumber\\
&\geq \pp{1-O(t^{-2})}e^{-o(\log t)-\delta(\gamma,\rho)\norm{\calP_t^\star}}\tilde{\pr}_1(\calE_1\cap\calE_2\vert\calE_0)\nonumber\\
&\geq \frac{\pp{1-O(t^{-2})}\log\frac{\alpha}{\beta}}{\log t}e^{-o(\log t)-\delta(\gamma,\rho)\norm{\calP_t^\star}}.\label{eqn:lowerBoundLessThenRho}
\end{align}
Combining \eqref{convTotal}, \eqref{eqn:MapProb2}, \eqref{eqn:lowerboundMismatchByE0E1E2}, \eqref{eqn:lowerboundE0}, and \eqref{eqn:lowerBoundLessThenRho}, we obtain
\begin{align}
\mathsf{E}(\calP^\star,\calQ) &= \liminf_{t\to\infty}-\frac{\log\mathsf{P}_{e,t}(\calP^\star,\calQ)}{\log t}\\
&\leq \liminf_{t\to\infty}-\frac{\log\pr_1\p{\calE_1\cap\calE_2\vert\calE_0}}{\log t}\\
&\leq \delta(\gamma,\rho)\liminf_{t\to\infty}\frac{\norm{\calP_t^\star}}{\log t} \\
&= \delta(\gamma,\rho)[(1+\gamma)\kappa(\gamma)],
\end{align}
as claimed.

\subsubsection{Upper Bound: $\rho\leq\rho_0$}\label{subsec:lowerBoundRho1_4}

For $\rho\leq\rho_0$ we use the fact that when the log-likelihood ratio is above $\log\alpha/\beta$, it has a downward drift. This implies that above $\log\alpha/\beta$, the walk cannot go beyond a certain value. Recall \eqref{eqn:MapProb2}. As before, in order to obtain a lower bound on \eqref{eqn:MapProb2} we define an event that implies that $\mathsf{R}_{t}<\beta/\alpha$. Now, when the log-likelihood ratio $\mathsf{J}_t\triangleq \log\mathsf{R}_t$ is above the line $\log\alpha/\beta$, using \eqref{eqn:mis_like1}--\eqref{eqn:mis_like2}, and  \eqref{eqn:likelihoodP_1}, we may write
\begin{align}
\mathsf{J}_s = \sum_{i=1}^s\xi_i,\label{eqn:randomWalk_1less0}
\end{align}
for $s\geq0$, where $\xi_i$'s are statistically independent random variables, and
\begin{align}
\pr_1\p{\xi_i=\log\frac{\beta}{\alpha}} = \frac{\beta}{\alpha+\beta}p_i,\label{eqn:randomwalk1_less}
\end{align}
and
\begin{align}
\pr_1\p{\xi_i=\log\frac{1-\frac{\beta}{\alpha+\beta}q_i}{1-\frac{\alpha}{\alpha+\beta}q_i}} = 1-\frac{\beta}{\alpha+\beta}p_i.\label{eqn:randomwalk2_less}
\end{align}
\begin{figure}[!t]
\begin{minipage}[b]{1.0\linewidth}
  \centering
	\centerline{\includegraphics[width=8cm,height = 5.5cm]{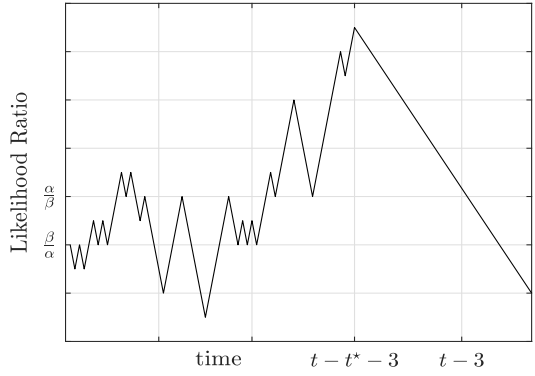}}
	\end{minipage}
	\caption{Illustration of a sample path of the random walk constructed in the lower bound.}
	\label{fig:walk_converse_bellow}
\end{figure}
Note that
\begin{align}
\bE_1\pp{\xi_i} &= \frac{(\gamma-1)\rho-\log\gamma}{1+\gamma}p_i+\Theta(p_i^2).\label{eqn:downwardRandomWalk_less}
\end{align}
Thus, we see that for $\rho\leq\rho_0$, we have $\bE_1\pp{\xi_i}\leq0$. We next show that with high probability $\max_{1\leq i\leq t}\mathsf{J}_i<\tau_0$, namely, the maximal value that the log-likelihood ratio can achieve is bounded by a certain constant $\tau_0$. Accordingly, using the same arguments as in the proof of Theorem~\ref{thm:2} this implies that only a finite number of timestamps are needed in order to drive log-likelihood ratio bellow $\log\alpha/\beta$. Specifically, as in the proof of Theorem~\ref{thm:2} it is suffice to assume that the last $t^\star+3$ revealers are such that their private information is $X_i=2$. Indeed, if for example, at time $\ell=t-(t^\star+3)$ the log-likelihood ratio $\mathsf{J}_\ell$ attained its maximal possible value $\tau_0$ (or $e^{\tau_0}$ for $\mathsf{R}_\ell$). Then, after $t^\star$ timestamps, i.e., at time $\ell=t-3$, the likelihood value is at most $\p{\beta/\alpha}^{t^\star}e^{\tau_0}$. Accordingly, if we set $t^\star = 1\vee(\frac{\tau_0}{\log\alpha/\beta}-1)$, then we get that the likelihood value is $\p{\beta/\alpha}^{t^\star}e^{\tau_0}\leq \alpha/\beta$, namely, bellow $\alpha/\beta$. Thus, in the worst case, at time $\ell=t-3$, the likelihood ratio value is in the interval $[\beta/\alpha,\alpha/\beta]$. In this interval, the MAP estimator outputs the private signal, namely, $Z_i=X_i=2$, and accordingly, the likelihood ratio is multiplied by $\beta/\alpha$. Therefore, the remaining $3$ timestamps simply insure that at time $t$ the likelihood ratio value is below $\beta/\alpha$, as required. To wit, if the likelihood value at time $t-3$ is $\alpha/\beta$, then at time $t$ it value will be $(\beta/\alpha)^2<\beta/\alpha$. Thus, by the above arguments, it is clear that we can lower bound the error probability as follows
\begin{align}
\pr\p{\mathsf{MAP}_{\calQ}(Z_1^{t},X_{t+1})\neq\theta} &\geq \pr_1\p{\mathsf{R}_{t}< \frac{\beta}{\alpha}}\\
&\geq \p{\frac{\beta}{\alpha+\beta}}^{3+t^\star}\pr_1\pp{\max_{1\leq s\leq t}\mathsf{J}_s\leq\tau_0}.\label{eqn:lowerBoundMaximamValue}
\end{align}
We next show that there exists a finite value of $\tau_0$ such that the probability term at the r.h.s. of \eqref{eqn:lowerBoundMaximamValue} is lower bounded by $1/2$. Thus, since $t^\star$ is finite, we obtain
\begin{align}
\mathsf{E}(\calP^\star,\calQ) &= \liminf_{t\to\infty}-\frac{\log\mathsf{P}_{e,t}(\calP^\star,\calQ)}{\log t}\\
&\leq \liminf_{t\to\infty}-\frac{\log\pp{\frac{1}{2}\p{\frac{\beta}{\alpha+\beta}}^{3+t^\star}}}{\log t}\\
&= 0,
\end{align}
as claimed. It is only left to prove that the probability term at the r.h.s. of \eqref{eqn:lowerBoundMaximamValue} is lower bounded by $1/2$. To this end, for any $\lambda\geq0$, we have
\begin{align}
\bE_1\pp{e^{\lambda\xi_i}}& = \frac{\beta}{\alpha+\beta}e^{\lambda\log\frac{\beta}{\alpha}}p_i+e^{\lambda\log\frac{1-\frac{\beta}{\alpha+\beta}q_i}{1-\frac{\alpha}{\alpha+\beta}q_i}}\p{1-\frac{\beta}{\alpha+\beta}p_i}\nonumber\\
& = \frac{\beta}{\alpha+\beta}\pp{1+\p{e^{\lambda\log\frac{\beta}{\alpha}}-1}}p_i+\pp{1+\lambda\rho\frac{\alpha-\beta}{\alpha+\beta}p_i+\Theta(p_i^2)}\p{1-\frac{\beta}{\alpha+\beta}p_i}\nonumber\\
& = 1+\pp{\p{e^{\lambda\log\frac{\beta}{\alpha}}-1}\frac{\beta}{\alpha+\beta}+\lambda\rho\frac{\alpha-\beta}{\alpha+\beta}}p_i+\Theta(p_i^2)\nonumber\\
& = 1+\pp{\p{e^{-\lambda\log\gamma}-1}\frac{1}{\gamma+1}+\lambda\rho\frac{\gamma-1}{\gamma+1}}p_i+\Theta(p_i^2).
\end{align}
Let us define the map,
\begin{align}
\varphi: \lambda \mapsto \p{e^{-\lambda\log\gamma}-1}\frac{1}{\gamma+1}+\lambda\rho\frac{\gamma-1}{\gamma+1}.
\end{align}
For $\lambda\ll1$, we have $\varphi(\lambda)= \frac{(\rho-\rho_0)(\gamma-1)}{\gamma+1}\lambda +O(\lambda^2)$, and since $\rho-\rho_0\leq0$, we may conclude that $\varphi(\cdot)$ has a negative derivative at $0$, hence its minimum, attained at $\lambda_0>0$, is strictly negative, namely, $\varphi(\lambda_0)<0$. Accordingly, due to statistical independence we may write,
\begin{align}
\bE_1\pp{e^{\lambda_0\mathsf{J}_s}} = \mathsf{C}_se^{\varphi(\lambda_0)\norm{\calP_s^\star}},
\end{align}
for a certain converging/bounded sequence $\{\mathsf{C}_s\}_{s\geq1}$. Next, define the random process \begin{align}
\mathsf{M}_s\triangleq\frac{\exp\p{\lambda_0 \mathsf{J}_s}}{\bE_1\pp{\exp\p{\lambda_0 \mathsf{J}_s}}},
\end{align}
for $s\geq1$. It is clear that $\{\mathsf{M}_s\}_{s\geq1}$ is a positive martingale. Thus, using Doob's martingale maximal inequality, we have for $\tau>0$,
\begin{align}
\pr_1\pp{\max_{1\leq s\leq t}\mathsf{M}_s\geq\tau}\leq\frac{\bE_1(\mathsf{M}_t)}{\tau} = \frac{1}{\tau},
\end{align}
which is equivalent to
\begin{align}
\pr_1\pp{\max_{1\leq s\leq t}\frac{e^{\lambda_0 \mathsf{J}_s}}{\mathsf{C}_se^{\varphi(\lambda_0)\norm{\calP_s^\star}}}\geq\tau}\leq \frac{1}{\tau}.
\end{align}
In particular, using the fact that $\varphi(\lambda_0)<0$, it is clear that the above implies that
\begin{align}
\pr_1\pp{\max_{1\leq s\leq t}e^{\lambda_0 \mathsf{J}_s}\geq\tau\cdot\max_{1\leq s\leq t}\mathsf{C}_s}\leq \frac{1}{\tau},
\end{align}
or,
\begin{align}
\pr_1\pp{\max_{1\leq s\leq t}\mathsf{J}_s\geq\frac{\log\pp{\tau\cdot\max_{1\leq s\leq t}\mathsf{C}_s}}{\lambda_0}}\leq \frac{1}{\tau}.
\end{align}
The above can be written also as follows
\begin{align}
\pr_1\pp{\max_{1\leq s\leq t}\mathsf{J}_s\geq\tau}\leq e^{-\lambda_0\tau}\cdot\max_{1\leq s\leq t}\mathsf{C}_s.
\end{align}
Therefore, taking $\tau>\tau_0\triangleq\frac{2\log\max_{1\leq s\leq t}\mathsf{C}_s}{\lambda_0}$, we have $\pr_1\pp{\max_{1\leq s\leq t}\mathsf{J}_s\geq\tau}<1/2$, as claimed.

\subsubsection{Upper Bound: $\rho_0\leq\rho\leq1$}\label{subsec:lowerBoundRho1_5}

For this regime we use similar arguments as in the previous subsection. The main difference here is that the log-likelihood ratio random walk now has a positive drift, and thus is unbounded. However, we claim that this unbounded value is small compared to $\log t$, and thus, the number of timestamps $t^\star$ needed to bring the random walk bellow $\log\frac{\beta}{\alpha}$ is small compared to $\log t$, and more importantly will not affect the learning rate. Specifically, recall \eqref{eqn:lowerBoundMaximamValue}. Taking $\tau_0 = \log^{3/4}t$, we have,
\begin{align}
\pr\p{\mathsf{MAP}_{\calQ}(Z_1^{t},X_{t+1})\neq\theta} &\geq \pr_1\p{\mathsf{R}_{t}< \frac{\beta}{\alpha}}\\
&\geq \p{\frac{\beta}{\alpha+\beta}}^{3+t^\star}\pr_1\pp{\max_{1\leq s\leq t}\mathsf{J}_s\leq\log^{3/4}t},\label{eqn:lowerBoundMaximamValue_2}
\end{align}
and since $t^\star = \Theta(\tau_0)$, we have
\begin{align}
\mathsf{E}(\calP^\star,\calQ) &= \liminf_{t\to\infty}-\frac{\log\mathsf{P}_{e,t}(\calP^\star,\calQ)}{\log t}\\
&\leq \liminf_{t\to\infty}-\frac{\log\pr_1\pp{\max_{1\leq s\leq t}\mathsf{J}_s\leq\log^{3/4}t}}{\log t}.\label{eqn:learning_rate_biggerRho0}
\end{align}
We next upper bound the r.h.s. of the above inequality. To this end, note that above $\log\alpha/\beta$, the log-likelihood ratio process $\mathsf{J}_s$ forms a random walk as in \eqref{eqn:randomWalk_1less0}--\eqref{eqn:randomwalk2_less}, now with a positive drift since $\rho\geq\rho_0$. We remove this drift by defining a new measure $\tilde{\pr}_1$, such that,
\begin{align}
\tilde{\pr}_1\p{\xi_i=\log\frac{\beta}{\alpha}} &= \nu_i,\\
\tilde{\pr}_1\p{\xi_i=\log\frac{1-\frac{\beta}{\alpha+\beta}q_i}{1-\frac{\alpha}{\alpha+\beta}q_i}} &= 1-\nu_i,
\end{align}
where 
\begin{align}
\nu_i\triangleq\frac{\log\frac{1-\frac{\beta}{\alpha+\beta}q_i}{1-\frac{\alpha}{\alpha+\beta}q_i}}{\log\p{\frac{\alpha}{\beta}\frac{1-\frac{\beta}{\alpha+\beta}q_i}{1-\frac{\alpha}{\alpha+\beta}q_i}}}.\label{eqn:nu_idef_less}
\end{align}
Now, under $\tilde{\pr}_1$, the random walk $\ppp{\mathsf{J}_{s}}_{s}$ is a martingale, and thus, using classical results on the tails of martingales (see, e.g., \cite{freedman1975,fan2015})
\begin{align}
\tilde{\pr}_1\p{\max_{1\leq s\leq t}\mathsf{J}_s\leq\log^{3/4}t}\geq 1-e^{-c\sqrt{\log t}},\label{eqn:lowerBoundTailMartingale}
\end{align}
for $t$ large enough. Next, performing a change of measure we may write
\begin{align}
&\pr_1\p{\max_{1\leq s\leq t}\mathsf{J}_s\leq\log^{3/4}t}= \tilde{\bE}_1\pp{\frac{\mathrm{d}\pr_1(\cdot)}{\mathrm{d}\tilde{\pr}_1(\cdot)}\mathds{1}\pp{\max_{1\leq s\leq t}\mathsf{J}_s\leq\log^{3/4}t}},\label{eqn:ChangeOfMeasure_less}
\end{align}
so we need to understand how the Radon-Nikodym derivative of $\pr_1(\cdot)$ w.r.t. $\tilde{\pr}_1(\cdot)$ behaves. Note that
\begin{align}
\frac{\mathrm{d}\pr_1(\cdot)}{\mathrm{d}\tilde{\pr}_1(\cdot)} &= \prod_{i=1}^t\left\{\frac{\frac{\beta}{\alpha+\beta}p_i^\star}{\nu_i}\mathds{1}\pp{\xi_i=\log\frac{\beta}{\alpha}}+\frac{1-\frac{\beta}{\alpha+\beta}p_i^\star}{1-\nu_i}\mathds{1}\pp{\xi_i=\log\frac{1-\frac{\beta}{\alpha+\beta}q_i}{1-\frac{\alpha}{\alpha+\beta}q_i}}\right\}.
\end{align}
We claim that each factor in the product can be lower bounded for some $C=C(\alpha,\beta)$ as follows
\begin{align}
&\frac{\frac{\beta}{\alpha+\beta}p_i^\star}{\nu_i}\mathds{1}\pp{\xi_i=\log\frac{\beta}{\alpha}}+\frac{1-\frac{\beta}{\alpha+\beta}p_i^\star}{1-\nu_i}\mathds{1}\pp{\xi_i=\log\frac{1-\frac{\beta}{\alpha+\beta}q_i}{1-\frac{\alpha}{\alpha+\beta}q_i}}\geq e^{\lambda^\star\xi_i} \tilde{K}_i(\xi_i),\label{eqn:RadonInequality_less}
\end{align}
where $\lambda^\star=\lambda_1^\star$ is defined in \eqref{eqn:lambdastarDef}, and
\begin{align}
\tilde{K}_i(\xi_i)&\triangleq e^{-\pp{\delta(\gamma,\rho)-\frac{\gamma-1}{\gamma+1}(1-\rho)}p_i^\star-C(p_i^\star)^2}\cdot\mathds{1}\pp{\xi_i=\log\frac{1-\frac{\beta}{\alpha+\beta}q_i}{1-\frac{\alpha}{\alpha+\beta}q_i}}\nonumber\\
&\quad\quad+e^{-\rho\p{\frac{1}{2}-\frac{\alpha-\beta}{(\alpha+\beta)\log\frac{\alpha}{\beta}}}p_i^\star-C(p_i^\star)^2}\cdot\mathds{1}\pp{\xi_i=\log\frac{\beta}{\alpha}}.
\end{align}
Indeed, this inequality can be checked for both potential values of $\xi$ by expanding the expressions in $p_i^\star$. Then, multiplying \eqref{eqn:RadonInequality_less} over all $i\in\{1,\ldots,t\}$, we obtain that
\begin{align}
\frac{\mathrm{d}\pr_1(\cdot)}{\mathrm{d}\tilde{\pr}_1(\cdot)}&\geq e^{\lambda^\star\mathsf{J}_{t}-C\sum_{i=1}^t(p_i^\star)^2}e^{-\pp{\delta(\gamma,\rho)-\frac{\gamma-1}{\gamma+1}(1-\rho)}\sum_{i\in\calV^c}p_i^\star}e^{-\rho\p{\frac{1}{2}-\frac{\alpha-\beta}{(\alpha+\beta)\log\frac{\alpha}{\beta}}}\sum_{i\in\calV}p_i^\star}\nonumber\\
&\geq e^{\lambda^\star\mathsf{J}_{t}-C'}e^{-\pp{\delta(\gamma,\rho)-\frac{\gamma-1}{\gamma+1}(1-\rho)}\norm{\calP_t^\star}}e^{-\rho\p{\frac{1}{2}-\frac{\alpha-\beta}{(\alpha+\beta)\log\frac{\alpha}{\beta}}}\sum_{i\in\calV}p_i^\star}\nonumber\\
&\geq e^{\lambda^\star\log\frac{\alpha}{\beta}-C'}e^{-\pp{\delta(\gamma,\rho)-\frac{\gamma-1}{\gamma+1}(1-\rho)}\norm{\calP_t^\star}} e^{-\rho\p{\frac{1}{2}-\frac{\alpha-\beta}{(\alpha+\beta)\log\frac{\alpha}{\beta}}}\sum_{i\in\calV}p_i^\star},\label{eqn:lowerbound_Radon2}
\end{align}
where $\calV\triangleq\ppp{i\geq 1:\xi_i=\log(\beta/\alpha)}$, $C'\triangleq C\sum_{i=1}^t(p_i^\star)^2$ is finite. As in \eqref{eqn:poissbinomTails}--\eqref{eqn:loglogBehaviour}, with probability at least $1-O(t^{-2})$, we have $\sum_{i\in\calV}p_i^\star= o(\log t)$. Therefore, combining this fact with \eqref{eqn:lowerBoundTailMartingale}, \eqref{eqn:ChangeOfMeasure_less}, and \eqref{eqn:lowerbound_Radon2}, we obtain
\begin{align}
\pr_1\p{\max_{1\leq s\leq t}\mathsf{J}_s\leq\log^{3/4}t}
&= \tilde{\bE}_1\pp{\frac{\mathrm{d}\pr_1(\cdot)}{\mathrm{d}\tilde{\pr}_1(\cdot)}\mathds{1}\pp{\max_{1\leq s\leq t}\mathsf{J}_s\leq\log^{3/4}t}}\nonumber\\
&\geq \tilde{\bE}_1\left[e^{\lambda^\star\log\frac{\alpha}{\beta}-C'}e^{-\pp{\delta(\gamma,\rho)-\frac{\gamma-1}{\gamma+1}(1-\rho)}\norm{\calP_t^\star}}\right.\nonumber\\
&\left.\quad\quad\quad\quad\cdot e^{-\rho\p{\frac{1}{2}-\frac{\alpha-\beta}{(\alpha+\beta)\log\frac{\alpha}{\beta}}}\sum_{i\in\calV}p_i^\star}\mathds{1}\pp{\max_{1\leq s\leq t}\mathsf{J}_s\leq\log^{3/4}t}\right]\nonumber\\
&\geq[1-O(t^{-2})]e^{-o(\log t)-\pp{\delta(\gamma,\rho)-\frac{\gamma-1}{\gamma+1}(1-\rho)}\norm{\calP_t^\star}}\tilde{\pr}_1\pp{\max_{1\leq s\leq t}\mathsf{J}_s\leq\log^{3/4}t}\nonumber\\
&\geq[1-O(t^{-2})](1-e^{-c\sqrt{\log t}})e^{-o(\log t)-\pp{\delta(\gamma,\rho)-\frac{\gamma-1}{\gamma+1}(1-\rho)}\norm{\calP_t^\star}}.\label{eqn:ChangeOfMeasure_less_last}
\end{align}
Finally, substituting \eqref{eqn:ChangeOfMeasure_less_last} in \eqref{eqn:learning_rate_biggerRho0}, we finally obtain
\begin{align}
\mathsf{E}(\calP^\star,\calQ) &= \liminf_{t\to\infty}-\frac{\log\mathsf{P}_{e,t}(\calP^\star,\calQ)}{\log t}\nonumber\\
&\leq \liminf_{t\to\infty}-\frac{\log\pr_1\pp{\max_{1\leq s\leq t}\mathsf{J}_s\leq\log^{3/4}t}}{\log t}\nonumber\\
&\leq \pp{\delta(\gamma,\rho)-\frac{\gamma-1}{\gamma+1}(1-\rho)}\liminf_{t\to\infty}\frac{\norm{\calP^\star_t}}{\log t}\nonumber\\
& = \pp{\delta(\gamma,\rho)-\frac{\gamma-1}{\gamma+1}(1-\rho)}(1+\gamma)\kappa(\gamma),
\end{align}
as claimed.

%%%%%%%%%%%%%%%%%%%%%%%%%%%%%%%%%%%%%%%%%%%%%%%%%%%%%%%%%
\subsection{Additional Proofs}\label{app:additional_Proofs}

\subsubsection{Proof of Theorem~\ref{thm:4}}
Note that the first two individuals follow their private signal, that is, $Z_i = X_i$, for $i=1,2$. Therefore, if if $X_1=X_2=2$, then it is clear that $\mathsf{R}_2 = (\beta/\alpha)^2$, which implies that the MAP
estimator outputs $2$ as its decision. Accordingly, it should be clear that if all future irrational players draw $2$ as their private information, then the
MAP estimator continues to output $2$. The above scenario gives a lower bound on the error probability. Specifically, let $\revt$ denote the set of revealers up to time $t$. It is clear that $\abs{\revt}$ follows a Poisson-Binomial distribution with mean $\mu = \norm{\calP_t}=o(\log t)$. Thus, for any $c>1$, using \eqref{eqn:poissbinomTails} we get
\begin{align}
\pr\pp{\abs{\revt}\geq c\norm{\calP_t}}\leq e^{-(c\log c-c+1)\norm{\calP_t}}.\label{eqn:revSizelogt}
\end{align}
Taking any $c$ such that $c\log c-c+1>0$, it is clear that the r.h.s. of \eqref{eqn:revSizelogt} is less than half, and
\begin{align}
\pr\p{\mathsf{MAP}_{\calQ}(Z_1^{t},X_{t+1})\neq\theta}&\geq \pr_1(X_1=2,X_2=2,X_i=2,\forall i\in\revt)\nonumber\\
&\geq \frac{1}{2}\p{\frac{\beta}{\alpha+\beta}}^2\pr_1\p{\left.\bigcap_{i\in\revt}\{X_i=2\}\right||\revt|\leq c\norm{\calP_t}}\nonumber\\
&\geq \frac{1}{2}\p{\frac{\beta}{\alpha+\beta}}^{2+c\norm{\calP_t}}.
\end{align}
Thus,
\begin{align}
\mathsf{E}(\calP,\calQ) &= \liminf_{t\to\infty}-\frac{\log\pr\p{\mathsf{MAP}_{\calQ}(Z_1^{t},X_{t+1})\neq\theta}}{\log t}\nonumber\\
&\leq [c\log(1+\gamma)]\cdot\liminf_{t\to\infty}\frac{\norm{\calP_t}}{\log t}=0,
\end{align}
as claimed.

%%%%%%%%%%%%%%%%%%%%%%%%%%%%%%%%%%%%%%%%%%%%%%%%%%%%%%%%%

\subsubsection{Proof of Theorem~\ref{thm:5}}

Since the proof of Theorem~\ref{thm:5} follows the steps of the proof of Theorem~\ref{thm:3} almost exactly, in this subsection we highlight the few technical differences only. Starting with the lower bounds, using the same steps as in Subsection~\ref{subsec:proof_upperBound}, one obtains the same upper bounds on the error probability as in \eqref{eqn:upperBounderrorgen1} and \eqref{eqn:upperBounderrorgen2}, for $1\leq\rho\leq\rho_1$ and $\rho_0\leq\rho\leq1$, respectively. In particular, for $1\leq\rho\leq\rho_1$ recall that
\begin{align}
\mathsf{P}_{e,t}(\calP,\calQ) &\leq \pp{C_0'e^{(1-\lambda_1^\star)C_3-\delta(\gamma,\rho)\norm{\calP_t}}+\frac{1}{t^2}}\cdot(1-p_t)+\frac{\beta}{\alpha+\beta} \cdot p_t.\label{eqn:upperBoundAppaga1}
\end{align}
Therefore, we get
\begin{align}
\mathsf{E}(\calP,\calQ)&=\liminf_{t\to\infty}-\frac{\log\mathsf{P}_{e,t}(\calP^\star,\calQ)}{\log t}\\
&\geq 1\wedge\pp{\delta(\alpha/\beta,\rho)\cdot\liminf_{t\to\infty}\frac{\norm{\calP_t}}{\log t}} \\
&= 1\wedge\pp{\mathsf{C_p}\cdot\delta(\gamma,\rho)},
\end{align}
as claimed. Similarly, using \eqref{eqn:upperBounderrorgen2}, we get that
\begin{align}
\mathsf{E}(\calP,\calQ)\geq1\wedge\pp{\mathsf{C_p}\cdot\p{\delta(\gamma,\rho)-\frac{\gamma-1}{\gamma+1}(1-\rho)}}, 
\end{align}
for $\rho_0\leq\rho_1$, as stated in Theorem~\ref{thm:5}. 

The upper bounds in Subsections~\ref{subsec:lowerBoundRho1}--\ref{subsec:lowerBoundRho1_5} remain the same as well. In fact, the only differences are in Subsections~\ref{subsec:lowerBoundRho1_3} and \ref{subsec:lowerBoundRho1_5}. Specifically, for $1\leq\rho\leq\rho_1$ the lower bound in \eqref{eqn:lowerBoundLessThenRho} still holds true. Then, recall that
\begin{align}
\mathsf{P}_{e,t}(\calP,\calQ) &= \pr\p{\mathsf{MAP}_{\calQ}(Z_1^{t-1},X_{t})\neq\theta}\cdot(1-p_t)+\pr(X_t\neq\theta) \cdot p_t\nonumber\\
&\geq \pr\p{\calE_0\cap\calE_1\cap\calE_2}\cdot(1-p_t)+\frac{\beta}{\alpha+\beta} p_t,\label{eqn:lowerBounderrorProbwithRev}
\end{align}
and thus combined with \eqref{eqn:lowerBoundLessThenRho}, we obtain
\begin{align}
\mathsf{E}(\calP,\calQ) &= \liminf_{t\to\infty}-\frac{\log\mathsf{P}_{e,t}(\calP,\calQ)}{\log t}\\
&\leq 1\wedge\liminf_{t\to\infty}-\frac{\log\pr_1\p{\calE_1\cap\calE_2\vert\calE_0}}{\log t}\\
&\leq 1\wedge\pp{\delta(\gamma,\rho)\cdot\liminf_{t\to\infty}\frac{\norm{\calP_t^\star}}{\log t}} \\
&= 1\wedge\pp{\mathsf{C_p}\cdot\delta(\gamma,\rho)},
\end{align}
as claimed. For $\rho_0\leq\rho\leq1$ we have a similar situation. Specifically, the lower bound in \eqref{eqn:ChangeOfMeasure_less_last} still hold true with $\calP^\star$ replaced by $\calP$. Also, recalling \eqref{eqn:lowerBoundMaximamValue_2} we have
\begin{align}
\mathsf{P}_{e,t}(\calP,\calQ) &= \pr\p{\mathsf{MAP}_{\calQ}(Z_1^{t-1},X_{t})\neq\theta}\cdot(1-p_t)+\pr(X_t\neq\theta) \cdot p_t\nonumber\\
&\geq \p{\frac{\beta}{\alpha+\beta}}^{3+t^\star}\pr_1\pp{\max_{1\leq s\leq t}\mathsf{J}_s\leq\log^{3/4}t}\cdot(1-p_t)+\frac{\beta}{\alpha+\beta} p_t.\label{eqn:lowerBounderrorProbwithRev2}
\end{align}
Thus, combining \eqref{eqn:ChangeOfMeasure_less_last} and \eqref{eqn:lowerBounderrorProbwithRev2}, we obtain
\begin{align}
\mathsf{E}(\calP,\calQ) &= \liminf_{t\to\infty}-\frac{\log\mathsf{P}_{e,t}(\calP,\calQ)}{\log t}\nonumber\\
&\leq 1\wedge\pp{\mathsf{C_p}\cdot\p{\delta(\gamma,\rho)-\frac{\gamma-1}{\gamma+1}(1-\rho)}},\nonumber
\end{align}
as stated in Theorem~\ref{thm:5}. 

%%%%%%%%%%%%%%%%%%%%%%%%%%%%%%%%%%%%%%%%%%%%%%%%%%%%%%%%%
\subsubsection{Proof of Theorem~\ref{thm:6}}

The proof of Theorem~\ref{thm:6} follows from two facts. First, recall that a trivial lower bound on the error probability is $\mathsf{P}_{e,t}(\calP,\calQ)\geq\frac{\beta}{\alpha+\beta}p_t$, which implies that $\mathsf{E}(\calP,\calQ)\leq -\lim_{t\to\infty}\log p_t/\log t$. We next show that if $\calQ$ is such that $\norm{\calQ_t}/\norm{\calP_t}\to\rho$, and $\rho_0<\rho<\rho_1$, then the above also lower bounds the learning rate. Indeed, as before, using the same steps as in Section~\ref{subsec:proof_upperBound}, we get the upper bounds in \eqref{eqn:upperBounderrorgen1} and \eqref{eqn:upperBounderrorgen2}, for $1\leq\rho\leq\rho_1$ and $\rho_0\leq\rho\leq1$, respectively. However, since in this case $\norm{\calP_t}=\omega(\log t)$ the terms in the squared brackets at the r.h.s. of \eqref{eqn:upperBounderrorgen1} and \eqref{eqn:upperBounderrorgen2} are negligible compared to the other $\frac{\beta}{\alpha+\beta}p_t$. This implies that \eqref{eqn:upperBounderrorgen1} and \eqref{eqn:upperBounderrorgen2} are dominated by $\frac{\beta}{\alpha+\beta}p_t$ and thus $\mathsf{E}(\calP,\calQ)\geq -\lim_{t\to\infty}\log p_t/\log t$, as well. Finally, it is left to show that $\mathsf{E}(\calP,\calQ)=0$ in the leftover cases, which follows from the same arguments as in Appendices~\ref{subsec:lowerBoundRho1} and \ref{subsec:lowerBoundRho1_4}, and therefore omitted.

%%%%%%%%%%%%%%%%%%%%%%%%%%%%%%%%%%%%%%%%%%%%%%%%%%%%%%%%

\subsection{Adversarial Model is Too Stringent }\label{app:too_string}
In this section we show that the error probability in \eqref{eqn:worst-case_error_def} associated with any estimator is lower bounded by a constant, and accordingly, the total number of errors in \eqref{eqn:total_error_def} is proportional to the number of players $\sN$. 

To this end, consider the set of revealers $\Pi_{\mathsf{N}} = [\sN-\sV+1:\sN]$ in \eqref{eqn:total_error_def}. This choice of $\Pi_{\mathsf{N}}$ corresponds to the case where all revealers appear at the end. Assume that $\sV=o(\sN)$, otherwise, $\mathsf{TE(V_N)}$ is trivially proportional to $\Theta(\sN)$. Then, since all first $\sN-\sV$ players are rational, with a positive probability a wrong cascade will occur. Indeed, this is just the classical herding experiment, proposed and studied in \cite{Anderson1,Anderson2} (see also \cite[Ch. 16]{David10}). In fact, each player $t\in[\sN-\sV]$ is wrong with probability at least $\frac{\beta^2}{(\alpha+\beta)^2} = (1+\gamma)^{-2}$, which is the probability that the decisions of the first two players are wrong (both draw marbles of minority type). Therefore, the number of errors in \eqref{eqn:total_error_def} satisfies  
\begin{align*}
\mathsf{TE}(\sV)&= \inf_{\hat{\theta}\in\hat{\Theta}}\sup_{\Pi_\mathsf{N}\subset[\mathsf{N}]:\;|\Pi_\mathsf{N}|=\mathsf{V_N}}\sum_{t=1}^\sN\mathsf{P}_{e,t}(\hat{\theta}_t,\Pi_\mathsf{N})\\
&\geq \frac{\sN-\sV}{(1+\gamma)^2},
\end{align*}
namely, of order $\Theta(\sN-\sV)$, which concludes the proof.

\section{Conclusion and Outlook}\label{app:future}

In this paper we have studied the effect of mismatch between players on information cascade, contrary to related works where full/partial mismatch was taken for granted. For the mismatch model considered in this paper we have identified when learning is possible and when it is not. Consequently, we demonstrated that the learning rate exhibits several surprising phase transitions. We hope our work has opened more doors than it closes. There are many questions for future work:
\begin{enumerate}
    \item It would be interesting to generalize our results to the case where more than two states are possible, each corresponding to multiple private signals. 
    \item In this paper we focus on . Studying the asymptotic learning rate and the 
    total number of wrong errors of information cascades over random graphs (e.g., Erd\H{o}s-R\'{e}nyi random graph, stochastic block models, etc.) is very interesting and of practical importance.
    \item Following our negative result on the worst-case model, studying minimax learning rates in adversarial models, by assuming a more \emph{structured} geometry for the set of revealers in order to avoid trivial rates is quite challenging and interesting. 
    \item It is important to check whether rational players that do not know $\calP$ can do better then just assuming some $\calQ$. In particular, devising a universal scheme that attains (or at least does not lose too much) the optimal learning rate for $\calP$, without knowing $\calP$, is an important question. A reasonable approach would be using the same $q_t = \Theta(t^{-1})$, and adapt the leading constant in some way. 
    \item As discussed in the introduction, it is well-documented in social learning literature that a fully rational model often places unreasonable computational demands on Bayesian players (e.g., \cite{mossel2017}), hence understanding the impact of simpler more efficient strategies is desirable. This situation can be partially captured by our model, since a sub-optimal mismatched MAP, e.g., a majority rule, can be employed by the players intentionally to reduce computational complexity. There are of course other computationally efficient strategies that cannot be covered by our mismatch MAP framework, but we hope that the results and techniques developed in our paper will prove useful in the analysis of other these strategies as well.
\end{enumerate}

%\subsection*{Acknowledgements}
%This research was supported by the European Research Council (ERC) under grant agreement 639573.

\bibliography{bibfile}{}
\bibliographystyle{abbrv}
\end{document}